\documentclass[preprint,12pt]{elsarticle}

\usepackage{amssymb}
\usepackage{amsmath}
\usepackage{verbatim}
\usepackage{tabularx}
\usepackage{booktabs}
\usepackage{url}
\usepackage{wrapfig, blindtext}
\usepackage{marginnote}
\usepackage{multicol}
\usepackage{multirow}
\usepackage{enumitem}
\usepackage{natbib}
\usepackage[table]{xcolor}
\usepackage{booktabs}
\usepackage{adjustbox}
\usepackage{subcaption}
\usepackage{tablefootnote}
\usepackage{array}
\usepackage{longtable}
\usepackage{multirow}
\usepackage{hyperref}
\usepackage{soul}

\usepackage{amsthm}
\theoremstyle{definition}
\newtheorem{definition}{Definition}[section]

\usepackage{color,soul}

\newcommand{\numproto}{$20$}
\newcommand{\dm}{\texttt{T2DM}}
\newcommand{{\ckd}}{\texttt{CKD}}

\journal{Artificial Intelligence In Medicine}

\begin{document}

\begin{frontmatter}

%% Title, authors and addresses

%% use the tnoteref command within \title for footnotes;
%% use the tnotetext command for theassociated footnote;
%% use the fnref command within \author or \address for footnotes;
%% use the fntext command for theassociated footnote;
%% use the corref command within \author for corresponding author footnotes;
%% use the cortext command for theassociated footnote;
%% use the ead command for the email address,
%% and the form \ead[url] for the home page:
%% \title{Title\tnoteref{label1}}
%% \tnotetext[label1]{}
%% \author{Name\corref{cor1}\fnref{label2}}
%% \ead{email address}
%% \ead[url]{home page}
%% \fntext[label2]{}
%% \cortext[cor1]{}
%% \affiliation{organization={},
%%             addressline={},
%%             city={},
%%             postcode={},
%%             state={},
%%             country={}}
%% \fntext[label3]{}

% \title{Supporting Contextualizations for User-centered Explainability in a Type-2 Diabetes Comorbidites Risk Prediction Setting}
\begin{comment}
\title{User-Centered Explainability to Improve Usefulness of AI Risk Prediction Models: A Case Study of Contexts from Medical Guidelines}
\end{comment}

\title{Informing Clinical Assessment by Contextualizing Post-Hoc Explanations of Risk Prediction Models in Type-2 Diabetes}
%perhaps a clinical guidelines case study.   (from dLLM)

%% use optional labels to link authors explicitly to addresses:
%% \author[label1,label2]{}
%% \affiliation[label1]{organization={},
%%             addressline={},
%%             city={},
%%             postcode={},
%%             state={},
%%             country={}}
%%
%% \affiliation[label2]{organization={},
%%             addressline={},
%%             city={},
%%             postcode={},
%%             state={},
%%             country={}}

\author[inst1]{Shruthi Chari}

\affiliation[inst1]{organization={Rensselaer Polytechnic Institute},%Department and Organization
            addressline={110 8th St}, 
            city={Troy},
            postcode={12180}, 
            state={NY},
            country={US}}

\author[inst1]{Prasant Acharya}
\author[inst1]{Daniel M. Gruen}
\author[inst2]{Olivia Zhang}
\author[inst2]{Elif K. Eyigoz}
\author[inst2]{Mohamed Ghalwash}
\author[inst1]{Oshani Seneviratne}
\author[inst3]{Fernando Suarez Saiz}
\author[inst2]{Pablo Meyer}
\author[inst2]{Prithwish Chakraborty}
\author[inst1]{Deborah L. McGuinness}

\affiliation[inst2]{organization={Center for Computational Health, IBM Research},%Department and Organization
            addressline={1101 Kitchawan Rd\\}, 
            city={Yorktown Heights},
            postcode={10598}, 
            state={NY},
            country={US}}

\affiliation[inst3]{organization={IBM Watson Health},%Department and Organization
            addressline={75 Binney St}, 
            city={Cambridge},
            postcode={02142}, 
            state={MA},
            country={US}}

\begin{abstract}
  Medical experts may use Artificial Intelligence (AI) systems with greater trust if these are supported by `contextual explanations' that let the practitioner connect system inferences to their context of use. However, their importance in improving model usage and understanding has not been extensively studied. Hence, we consider a comorbidity risk prediction scenario and focus on contexts regarding \textit{the patients' clinical state, AI predictions about their risk of complications, and algorithmic explanations supporting the predictions}. We explore how relevant information for such dimensions can be extracted from Medical guidelines to answer typical questions from clinical practitioners. We identify this as a question answering (QA) task and employ several state-of-the-art Large Language Models (LLM) to present contexts around risk prediction model inferences and evaluate their acceptability. Finally, we study the benefits of contextual explanations by building an \textit{end-to-end} AI pipeline including data cohorting, AI risk modeling, post-hoc model explanations, and prototyped a visual dashboard to present the combined insights from different context dimensions and data sources, while predicting and identifying the drivers of risk of Chronic Kidney Disease ({\ckd}) - a common type-2 diabetes ({\dm}) comorbidity. All of these steps were performed in deep engagement with medical experts, including a final evaluation of the dashboard results by an expert medical panel. We show that LLMs, in particular BERT and SciBERT, can be readily deployed to extract some relevant explanations to support clinical usage. To understand the value-add of the contextual explanations, the expert panel evaluated these regarding actionable insights in the relevant clinical setting. Overall, our paper is one of the first end-to-end analyses identifying the feasibility and benefits of contextual explanations in a real-world clinical use case. Our findings can help improve clinicians' usage of AI models.

\end{abstract}

%SC Adding TOC - will remove this later 
%\tableofcontents      
%\listoffigures  
%\listoftables          

%%Graphical abstract - Uncommenting as this was confirmed to be not required
% \begin{graphicalabstract}
% \includegraphics{grabs}
% \end{graphicalabstract}

%%Research highlights
%question does the journal ask for research highlights?  and will these be published at the beginning of the paper?  and how are these highlights supposed to connect/overlap with the abstract?
%SC Research higlights are meant to be a point wise summary of the contributions 

\begin{highlights}
\item Generate contextual explanations for a comorbidity risk prediction setting.
\item Explanations tie predictions and posthoc features to evidence from the context of use.
\item Explanations extracted from guidelines using LLMs and knowledge-augmentations (KAs).
\item Evaluate 5 clinical LLMs and KAs on an entire dataset and divide them by diseases.
\item An expert panel found value in providing such contextual explanations.
%https://ibm.box.com/s/pdp39lp2u032ic4r7rvciwy8yg8elk7c - higlights here
\end{highlights}

% Above worded to match Dan's suggestions from Box
% Importance of knowing what the context of use is for determining what explanations are needed
% Itemizing three dimensions around which to provide explanations, informed by clinical users
% Mechanisms to provide those explanations
% Sample UX used as a probe to collect feedback
% Lessons from that feedback

\begin{keyword}
%% keywords here, in the form: keyword \sep keyword
User-driven, Clinical Explainability \sep Contextual Explanations \sep Question-Answering Approach \sep Type-2 Diabetes Comorbidity Risk Prediction
%question - are these keywords taken from a list that the journal provides?  the keywords have explanation or explainablity 3 times
%% PACS codes here, in the form: \PACS code \sep code
\PACS 0000 \sep 1111
%% MSC codes here, in the form: \MSC code \sep code
%% or \MSC[2008] code \sep code (2000 is the default)
\MSC 0000 \sep 1111
\end{keyword}

\end{frontmatter}

%% \linenumbers

%% main text
\section{Introduction}
\label{sec:introduction}
%different uses of context - context of uses, different sources of context, and there is a relationship that is the contextualization ; explanations refer to these sources of context and need to be relevant in the context of use 

Artificial Intelligence (AI) and Machine Learning (ML) have been applied to the medical and healthcare domains for decades~\cite{shortliffe1974mycin, briganti2020artificial} but their adoption has been slow due to various aspects including the need for explaining the black-box nature of such methods. AI explainability (XAI) has tried to provide a rationale for model predictions so that subject matter experts (SMEs) can interpret their results~\cite{swartout1991explanations, gunning2017explainable, Chari2020FoundationsFE}. Studies on XAI have shown that users in different contexts require explanations that match their different levels of expertise and focused on their particular goals and needs~\cite{dey2022human, GHASSEMI2021e745, chakraborty2020tutorial, chari2020explanation}. Hence, there often is no single solution for XAI pipelines~\cite{gilpin2018explaining, arya2019one, chari2020explanation}, but processes can be followed for catering to the specific needs of the use case and the intended users~\cite{liao2020questioning, wang2019designing}. The impact of AI in patient-facing applications will increase the need not only for XAI but also for contexts that subject matter experts (SMEs) in the clinical domain are familiar with~\cite{GHASSEMI2021e745, doshi2017towards, tonekaboni2019clinicians}.

%SC add from our interactions - feature importances are not enough 
%SC add subsection on coined term - trustworthy and more applicable 

%SC do we need a ending sentence here 
%be careful about the usage of context 

%SC where is the gap for clinicians, say "In this work we provide explanations to bridge the gap"
%SC add plug to contextual explanations from book chapters 
% More specifically, the risk prediction setting of a chronic disease like type-2 diabetes ({\dm}) combines of all the previously mentioned traits. 

\begin{definition}[Explanation]
An account of the system, its workings, the  \textit{implicit and explicit} knowledge it uses to arrive at conclusions in general and the specific decision at hand, that is \textit{sensitive} to the end-user's \textit{understanding, context, and current needs}.~\cite{Chari2020DirectionsFE}
\end{definition}

% \textit{Contextual explanations:} 
\begin{definition}[Contextual Explanation]
  Explanations that contain context, are often explicit information ~\cite{lieberman2000out} to characterize the situation of (an) entity(ies), wherein ``an entity is a person, place, or object that is considered relevant to the interaction between a user and an application''~\cite{dey1998cyberdesk}. 
\end{definition}

In recent years, efforts to describe and formalize  explanations~\cite{chari2019making,chari2020explanation} have identified various dimensions and types for it. Specifically, `contextual explanations'~\cite{challener2019proliferation, chari2020explanation} hold great promise to satisfy aforementioned clinical needs and can improve the adoption of AI methods among clinical workflows. Risk prediction is one of the most important tasks in clinical decision making, and an increasingly important in view of the move toward personalized medicine.~\cite{videha2021adoption, banning2008review}. To interpret risk scores, clinicians often consult evidence from different levels of the scientific pyramid~\cite{rosner2012evidence} to lookup associations that might impact the patient's treatment or future trajectory. For example, questions like those in Tab. \ref{tab:samplecontextquestions}, are often asked by clinicians when they are trying to understand or use AI model predictions in their practice. Additional contextual information, such as answers to these questions, can help clinicians interpret and trust predictions to take actions. 
However, current work in risk prediction has often narrowly focused on improving model's accuracy, ignoring the aforementioned needs.
Interestingly, several researchers have posited contextual explanations~\cite{lakkaraju2022rethinking, framling2020decision, chari2020explanation}, that go beyond post-hoc model explanations to frame the predictions in the context of the applied setting and decisions being made. However, the feasibility of extracting such contextual explanations and the added benefit in an end-to-end setting of clinical relevance has not been studied and forms the focus of this paper. Specifically, we consider how to derive and support contextual explanations from authoritative domain knowledge sources, not already considered by prediction models, that clinicians would typically use to reason through decisions presented to them when dealing with recommendations from learning health systems.
% In this manuscript, we present one such, end-end multi-method approach to support \textit{contextual explanations} in clinical settings, particularly, in the risk prediction of comorbodities of a chronic disease like Diabetes. 
% Additionally, since, the need for explanations changes based on context, we attempt to generate these explanations to reference and connect to additional information that is involved in or relevant to that context and we study whether such contextual explanations are useful.
% \todotip[SC]{see if this sentence is still needed}
%SC commenting

\begin{table}[!htbp]
\centering
\caption{Questions that could be asked in clinical use cases around model explanations / predictions, and which can benefit from contextual explanations in the context of use.} 
\small
   \begin{tabular}{l}
%   \caption{Questions that could be asked in clinical use cases around model explanations / predictions, and which can benefit from contextual explanations in the context of use.} 
    \toprule
    Sample Question \\
    \midrule
    %One-class SVM & & \\
     \multicolumn{1}{m{\columnwidth}}%{m{6cm}}
     {What treatment can be suggested for this patient who has an increased risk of cardiovascular disease?} \\ \midrule
    \multicolumn{1}{m{\columnwidth}}%{m{6cm}}
    {What other conditions does this patient have that might impact this decision?} \\ \midrule
    \multicolumn{1}{m{\columnwidth}}%{m{6cm}}
    {What was the patient's A1C value when this prediction was made?} \\ \midrule
    \multicolumn{1}{m{\columnwidth}}%{m{6cm}}
    {Why are you telling me that this risk is important?} \\
    \bottomrule
\end{tabular}
\label{tab:samplecontextquestions}
\end{table}

Working closely with medical experts, we identify three specific focus areas on contexts such as \textit{the patients' clinical state, AI predictions about their risk of complications, and algorithmic explanations supporting the predictions}. 
Multiple data sources are required to extract such contextual explanations, including patient medical records, AI model predictions, and authoritative information around clinical facts and best practices. Medical guidelines are one of the most trusted authoritative sources of information and can provide the required additional context. Here, we thus study the feasibility of extracting answers from guidelines to typical questions from clinical practitioners to satisfy their explainability needs. We can identify this problem to be a question answering (QA) task. 
In the natural language processing (NLP) domain, the efficacy and ready-availability of state-of-the art (SOTA) deep learning based QA modules, have rendered such QA problems solvable and is being increasingly productivized. In this paper, we thus aim to extract such contextual explanation using SOTA LLM methods. Especially, we aim to study the following questions relative to contextual explanations:
% Specifically, design a case-study around  we attempt to answer the below questions about contextual explanations including their:
%\begin{itemize}
\par \noindent $\bullet$ \ul{Feasibility of extracting and generating contextual explanations from authoritative sources}: \textit{can we reliably extract contextual explanations from medical guidelines using state-of-the-art QA models? Can knowledge augmentation improve the QA performance?} We use a suite of readily available QA language models, with and without knowledge augmentations, and compare against manually annotated answers to evaluate the extracted contexts from authoritative clinical sources to explain decisions of post-hoc model explainers and risk prediction models.

    % \begin{itemize}
        % \item Are question-answer (QA) technologies performant? We evaluate the abilities of off-the shelf state of the art (SOTA) language models with some knowledge augmentations, to support our hypothesis of providing contexts from authoritative clinical sources to explain decisions of post-hoc model explainers and risk prediction models.
    % \end{itemize}
\par \noindent $\bullet$\ul{Understanding the added benefit of the derived contexts}: \textit{does the derived contextual explanations improve model usage by clinicians?} We evaluate usefulness of these contexts from two perspectives: (a) user persona needs and (b) model accuracy. Particularly, we discuss themes that emerged from our conversations with clinicians to understand if clinical contexts can better support model explanations and what more is desired of contextual explanations.
        % \item Model accuracies and detailed evaluations that point to the strength of SOTA methods to provide contexts.
    % \end{itemize}
\par \noindent $\bullet$ \ul{Practical considerations in a clinical workflow}:
\textit{what considerations and challenges might we face in implementing support for derived contexts in a setting of clinical relevance?}
To conduct our study, we developed an end-to-end system including (i) data cohorting, (ii) AI models for risk prediction, (iii) post-hoc explainers to identify driver of risk, and (iii) a prototype dashboard to present the combined insights from the contextual explanations. 
    Specifically, as a case-study, we considered the problem of predicting and identifying the drivers of risk of chronic kidney disease ({\ckd}) - a common type-2 diabetes ({\dm}) comorbidity and extracted contextual explanations for 175 questions of 5 different types. An expert panel of $4$ medical experts evaluated these explanations for $20$ prototypical patients. Our end-to-end system enabled us to identify practical steps in creating such a pipeline and the steps needed to generalize this for other clinical settings. Also, this enabled us to answer the aforementioned two questions and derive a holistic understanding supporting contextual explanations in a risk prediction setting. We further identify scenarios within this clinical setting where contextual explanations would be most useful and discuss how they would fit into the clinical workflow.
%\end{itemize}

The rest of the paper is organized as follows. First, we provide a brief description of the use case, the data sources, and the motivation and background for contextual explanations (Sec.~\ref{sec:background}). Next, we provide an overview of our methodology including the end-to-end AI pipeline and prototype dashboard used to conduct the experiments in (Sec.~\ref{sec:methods}).
% of our end-end implementation to support explanations in the context of entities related to risk prediction settings.  
In Sec.~\ref{sec:results}, we present comprehensive evaluations including quantitative performance of QA language models as well as qualitative analysis of extracted contextual explanations via a thematic analysis of expert panel sessions.
% model performance numbers of the risk prediction models for diabetes comorbidities and post-hoc explainers, 
% report themes from our expert panel sessions, and provide accuracy numbers and coverage statistics for the guideline QA module. 
We present a detailed analysis of these results in Sec.~\ref{sec:discussion} and answer the aforementioned questions of interest.
% themes that emerged from the expert panel interviews and do so with an eye towards how well they can be/are supported by the contextualizations we support and the domain knowledge sources we consider. 
Finally, in Sec.~\ref{sec:relatedwork}, we present a summary of other related works and contrast our unique contributions, 
% our contextual explanation methods against published studies that provide either knowledge-enabled or contextual explanations or describe approaches to query clinical literature 
and conclude with general take-aways, including opportunities for future research in Sec~\ref{sec:conclusion}.

% %SC End with a high-level summary of what is said in the paper.

\section{Motivation and Background}
\label{sec:background}
In this section, we provide details on several assumptions and considerations that will be used to describe our methodology and will be crucial in analyzing the experimental results.
Here, we describe the content we will use to support user-centered contextual explainability in chronic disease - comorbidity risk prediction settings. In Sec. \ref{sec:use-case}, we present a high-level overview of the selected use-case.
%summary of why the risk prediction of {\ckd}, a comorbidity of a chronic disease like {\dm} is an important problem. Then, in Sec. \ref{sec:setting}, we summarize our discussions with a clinical expert to describe where a chronic disease, comorbid risk prediction tool like ours would fit into the clinical workflow. Upon that, 
in Sec. \ref{sec:contextualizations}, we introduce the entities along which we extract contextual explanations (or contextualized entities), which could provide additional information to help clinicians interpret the risk prediction scores and the factors influencing the scores in clinical settings. 
Finally, in Sec.~\ref{sec:data-sources}, we provide an overview of the datasets used for our study. In particualr, we describe clinical practice guidelines (CPGs) in Sec. \ref{sec:cpg}. CPGs are considered to be at the highest level of the evidence-based pyramid~\cite{rosner2012evidence} and is our selected source to derive high-quality clinical context. 
% and describe the typical structure of CPGs, 
%We use the details we mention here to build towards a description of our methods to provide context from multiple sources around three entities of interest in the risk prediction of {\ckd} among {\dm} patients.
% Finally, in Sec. \ref{sec:problem-statement}, we leverage these descriptions to present our problem statement, which forms the basis for our multi-method approach to extract contextualizations in our use case and serves as the hypothesis for the expert panel sessions we conducted to validate our methods and results. 

\subsection{Use Case}
 \label{sec:use-case}
 % \todotip[PC]{lead}
% \explannote[PC]{done}
\begin{figure*}[!htbp]
    \centering
    \includegraphics[width=\linewidth]{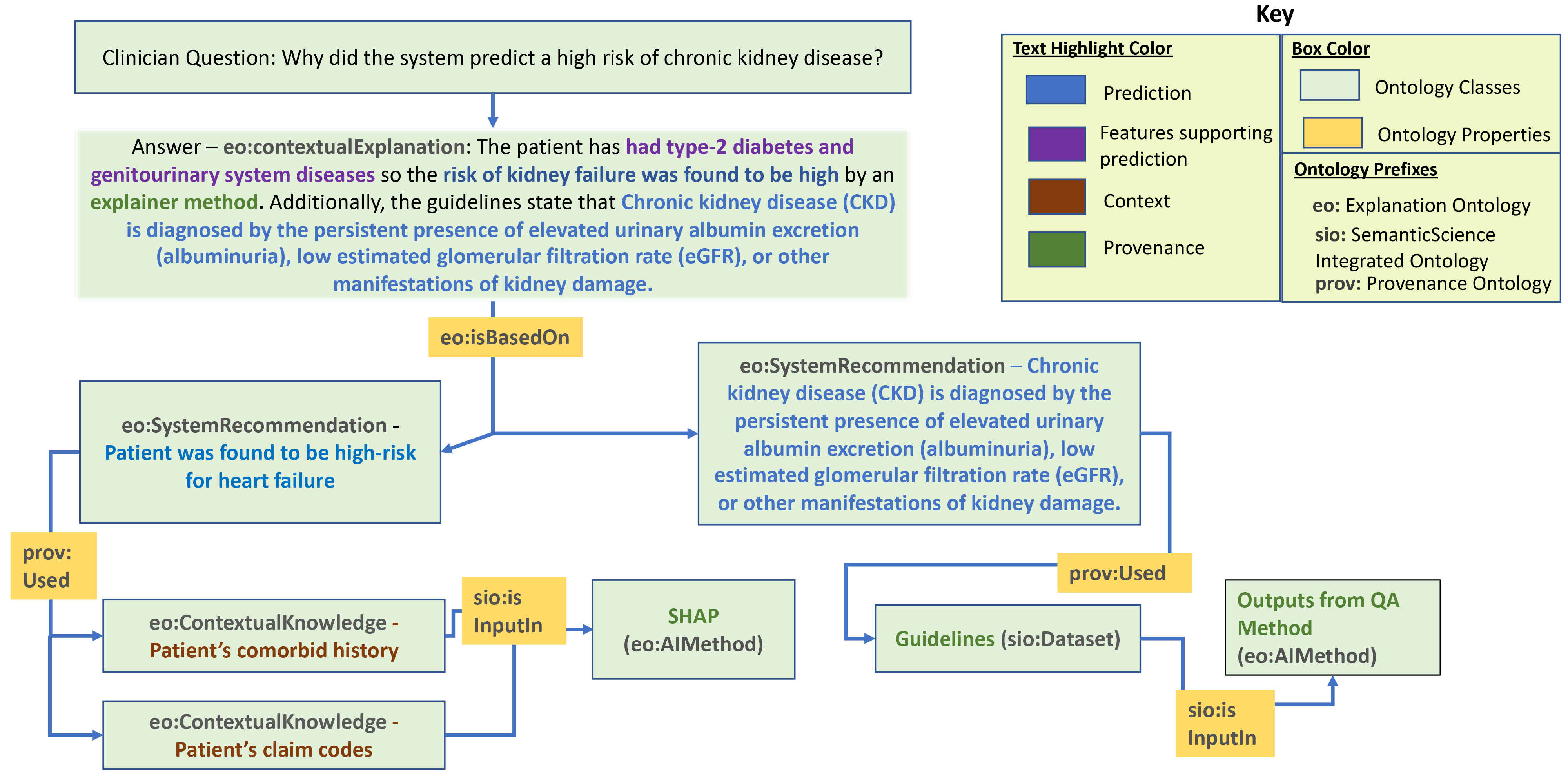}
        \caption{Trace of dependencies of a contextual explanation example on system outputs from post-hoc explainer models and question-answering methods. In this example, we have used ontology classes and properties to annotate the data, but in this paper we mainly focus on a multi-method approach to support such contextual explanations.}
        \label{fig:usecaseexample}
\end{figure*}

%  \todotip[PC]{lead}
AI models promises to help clinicians by providing tools for improved decision making. Risk prediction of patients is one of the key steps in a clinical decision scenario and models for such use cases can be consumed by a broad spectrum of clinicians with differing roles and experience, e.g. a specialist vs a primary-care physician. Depending on their roles, the needs, and thereby the desired functionality from a risk prediction contextualization standpoint, can be different. We worked closely with a clinical expert to
to understand the clinical use-case and determine the context of AI tools. 
Crucially, we aimed to form an understanding of the unmet needs and identify relevant contexts that can benefit clinicans. We can further motivate this via Fig.~\ref{fig:usecaseexample} which shows an example question posited by a clinician while consuming the ouputs of a risk prediction model. In this case, the relevant response can be identified via contextual explanations~\cite{chari2020explanation} that is generated from multiple sources and via multiple extracted contexts.
Our aim, was to thus scope the relevant contexts that will be the focus of our study. 
We followed a sequence of %standard 
user-centric research principles~\cite{liao2020questioning}, to
(1) define the scope of our tool's capabilities, (2) identify the end-user/target persona who would most benefit from our tool, and (3) scope the most relevant contexts. %and realistic contexts and usage scenarios for our tool.
Through our interviews, we identified primary-care physicians (PCP), especially those with lesser years of experience, to be the persona who may most benefit from such contextualizations. We describe the covered context types in Section~\ref{sec:contextualizations}. 
Furthermore, to study our stated problem in a real-world setup, we identified the problem of 
%Based on these understandings, we formalize our use case to \textit{provide contextual explanations to a PCP around the 
risk prediction of {\ckd} among new {\dm} patients at their first diagnosis.

This is motivated by the fact that diabetes is one of the top five chronic diseases affecting the adult population in the US~\cite{cdcT2D}. 
Diabetes management involves monitoring for and treating %the 
related comorbid conditions. 
Effective and timely prediction of such conditions can lead to an overall improvement in the quality of care and thus evaluating the impact of AI models in improving clinical decision workflow can have tremendous real-world impact.
%We focus on {\dm} and use machine learning (ML) models to predict the risk of developing certain {\dm} comorbid complications. A
Especially, we focus on {\ckd}, a commonly occurring micro-vascular complication of {\dm} and one of the leading causes of death in the US~\cite{cdcCKD}, with an estimated $37$ million cases in the US (who are mostly undiagnosed) and cost medicare in 2018 $81.1B$, and end stage renal disease an additional $36.6B$. Typically, actions to prevent onset of {\ckd} among {\dm} patients revolve around proper disease control,  including close disease monitoring, proper treatment adherence, and patient education. Incorporating accurate risk prediction of {\ckd} in the clinical workflow can lead to more timely actions, potentially delaying the onset of {\ckd}, and in some cases, preventing its progression.
While such predictions could be of use along various time-points of the patients' {\dm} prognosis, in this paper, we predict the risk of developing {\ckd} within 360 days of {\dm} onset. 
Under this use case, we explore strategies to provide context around interventions for particular patients, and explain their {\dm} state and individual risk factors.

 %\input{tex/use_case}

%\subsection{Clinical Setting} 
%\label{sec:setting}
%\input{tex/clinical_scenario}

\subsection{Selected Contextual Entities of Interest} \label{sec:contextualizations}
To support the goal of providing user-centered, clinically relevant, and contextual explanations, 
in consultation with a medical expert on our team who is also a co-author, we identify three entities of interest to provide contextual explanations around predicting the risk of {\ckd} among {\dm} patients. Fig.~\ref{fig:usecaseexample} shows an example of contextual explanation that can answer a clinician's question around patient management. It can be seen that such explanations are usually composed of multiple entities and from multiple sources. In general, we identified and subsequently focused on extracting the following contexts:

\par \noindent $\bullet$ \ul{Contextualizing the patient} by connecting their clinical history and indicators to treatments typically recommended for such patients, according to CPGs.
    %SC previous point: doesn't include risk prediction per se
\par \noindent $\bullet$ \ul{Contextualizing risk predictions for the patient} in terms of the prediction's impact on decisions, based on general norms of practice concerning potential complications, as evident from guidelines and other domain knowledge, including medical ontologies.
   %SC mention expanding chapters in discussion
\par \noindent $\bullet$  \ul{Contextualizing details of algorithmic, post-hoc explanations}, such as connecting features that were the most important to other information based on their potential medical significance, such as through connections to physiological pathways and CPGs.

\begin{figure*}[!htbp]
    \centering
    \includegraphics[width=\linewidth]{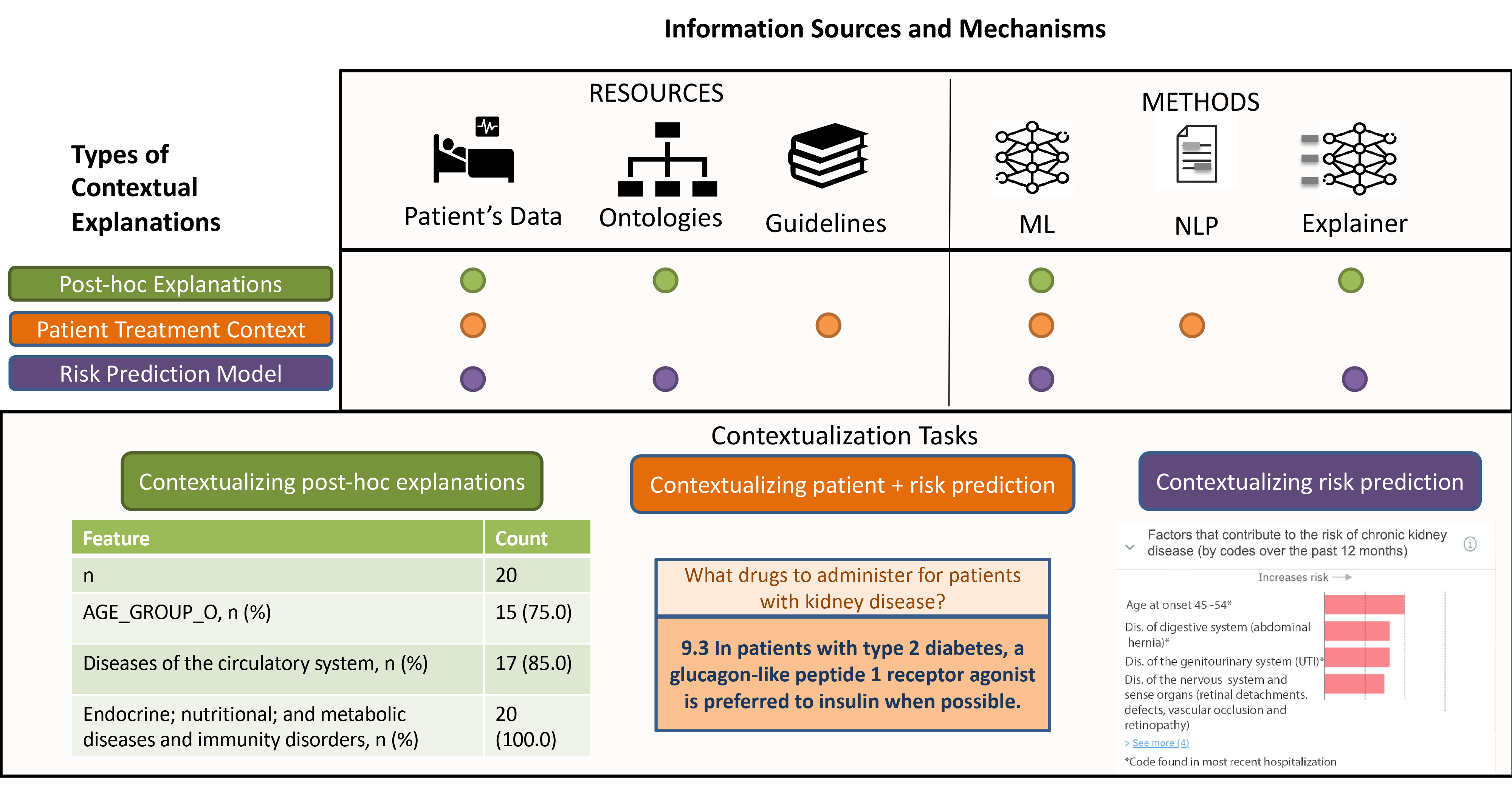}
        \caption{Different types of contextualizations supported by methods, that help provide additional context around patients, their risk predictions and features contributing to risk, via connections to different knowledge sources including patient data, medical ontologies and guidelines.}
        \label{fig:contextualizations}
\end{figure*}

In Fig. \ref{fig:contextualizations}, some examples of contexts that we support around the three entities of interest in the risk prediction setting can be seen. Also seen in the figure are the pathways in which answers providing context could borrow from different domain knowledge sources and methods. For example, the answer to the question, ``What drugs to administer for chronic kidney disease?'' provides context around the patient and risk prediction, borrows both from guidelines and patient data, and is supported by the risk prediction and natural language modules which we describe in Sec. \ref{sec:methods}.  

\subsection{Data Sources}
\label{sec:data-sources}
To conduct our real-world study, we focus on two specific sources of data as described below.
\subsubsection{Patient Data} \label{sec:patient-data}
We conduct our analysis on and retrieve patient data from the claims sub-component of the Limited IBM MarketScan Explorys Claims-EMR Data Set (LCED), covering both administrative claims and EHR data of over $5$ million commercially insured patients between 2013 and 2017.
Medical diagnoses are encoded %in the database 
using International Classification of Diseases (ICD) codes. 
We selected only those {\dm} patients (with ICD9 codes 250.*0, 250.*2, 362.0, and ICD10 code E11)  that satisfied the following criteria as our cohort.
 Only {\dm} patients with the following criteria are included:% in the study:
\begin{itemize}
  \item have had two or more visits with {\dm} diagnosis,
  \item  were enrolled continuously for $12$ months prior to %initial 
{\dm} diagnosis,% with {\dm}. 
  \item  number of visits for {\dm}  is greater than those for other forms of diabetes such as \texttt{T1D}, and
  \item  age at the initial {\dm} diagnosis is between 19-64 years.
\end{itemize}

Among {\dm} patients, we use the first diagnosis of chronic kidney disease ({\ckd}) (ICD10 N18 or ICD9 585.*, 403.*) after the initial diagnosis of {\dm} as the outcome to predict. %In this paper, a
At the time of the first {\dm} diagnosis, we predict the risk of the patient developing {\ckd} within $1$ year using Clinical Classifications Software (CCS) codes, age group, and sex as features for the predictive model.

\subsubsection{Clinical Practice Guidelines} \label{sec:cpg}
Clinical Practice Guidelines are position statements published by a board of experts in different disease areas~\cite{murad2017clinical}. These guidelines are updated often, latest summaries of updated evidence in the disease areas, and follow the highest standards of evidence appraisal (e.g., Grading of Recommendations, Assessment, Development and Evaluations (GRADE) evidence schemes~\footnote{https://www.gradeworkinggroup.org}). Further, the guidelines are written to be comprehensive sources covering different aspects of treatment, management, and assessment of the disease and are often regarded as first-line lookup sources for clinicians and primary care physicians~\cite{graham2011trustworthy, murad2017clinical}. Given their comprehensive and updated nature, CPGs provide a great resource for providing clinical contexts in various clinical settings. We utilize the 2021 edition of the American Diabetes Association (ADA) Standards of Care guidelines for our experiments.

\section{Methods}
\label{sec:methods}
\begin{figure*}[!htbp]
    \centering
    \includegraphics[width=\linewidth]{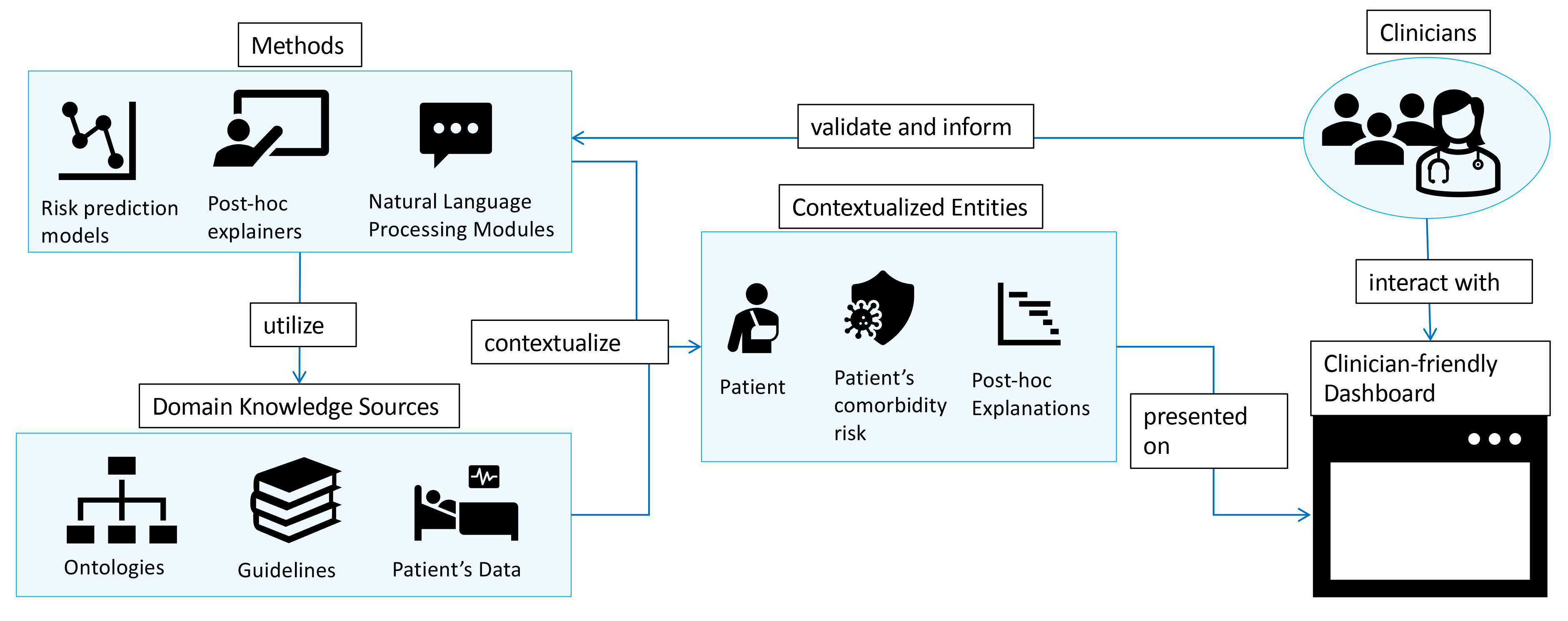}
        \caption{Overall view of the different methods in our pipeline and how they interact to provide risk prediction scores, factors contributing to the risk and contexts around the patient, their predicted risk and the factors contributing to the risk.}
        \label{fig:overalLLMethodsummary}
\end{figure*}

 To study the problem of risk prediction of {\ckd} among {\dm} patients, we created an end-to-end AI enabled system. Fig. \ref{fig:overalLLMethodsummary} shows a conceptual overview of the components of this system. In general, to extract contextual explanations around our three identified entities of interest, we used a number of components including risk prediction models, post-hoc explanation models, and our multi-method, question-answering approach to provide context. %We describe each of these methods in this section. 
 %These methods and their interactions with our many supported data sources can be seen in Fig. \ref{fig:overalLLMethodsummary}. 
Crucially, to analyze the importance of the supported contextual explanations, we prototyped a clinical-friendly dashboard and ran qualitative analysis. 
%We describe an overview of the same in Sec~\ref{sec:dashboard}.
%present our risk predictions, supporting patient data, model provenances, and supported contextual explanations on a clinical-friendly dashboard (Sec. \ref{sec:dashboard}).
%
In this section, we provide high-level details of some of the key components involved in the process.

\iffalse
Hence, we center our methods to provide context around:
\begin{enumerate}
    \item The identification of three entities to provide clinically relevant contexts in the risk prediction setting. 
    %SC developing mechanisms around contexts 
    \item The information extraction of relevant contexts around these entities, specifically in the {\dm}, {\ckd} risk prediction setting.
    %SC add a sub-point on the guidelines 
    \item Supporting the combination of these contexts, using our novel unsupervised, knowledge-guided language model approach, from different evidence-based sources, including authoritative sources like guidelines and well-established medical ontologies, and patient data via a question-answering approach. 
    %SC change the above wording 
    \item Finally, supporting the evaluation of the usefulness of these contexts from perspectives covering user persona needs, content coverage metrics, and model accuracies. 
\end{enumerate}
\fi

\subsection{Risk Prediction Models}
    \label{ssub:risk_model}
    In the first step of our pipeline, we build risk prediction models from the constructed cohort and the aforementioned use-case (Sec.~\ref{sec:use-case}). In particular, we train a suite of machine learning models (ML), including both classical and deep-learning models, and select the best performing one based on the highest predictive accuracy and other appropriate metrics for the use case, such as favoring models with a higher recall. We used DPM 360~\cite{dpm360}, an open-source, reusable, disease progression model training package, we compared a suite of classification models on the patients' demographic and diagnosis history to predict future complications. In this paper, we only used the demographic and diagnostic features to model risk. Furthermore, to handle the temporal features, for some of our models, such as Logistic Regression (LR) and Multi-layer perceptron (MLP), we used temporally aggregated features (summation). 
We also compared two state-of-the art Recurrent Neural Networks (RNN) where temporal history can be handled in a more natural manner, viz., Long-Short Term Memory (LSTM) and Gated Recurrent Units (GRU). All model implementations are available via DPM360 including classical ML models (backed by scikit-learn) and deep learning models (custom built for DPM).
In this paper, we split the data according to a train-validation-test split (70-10-20). Using the best performing models on the validation set, we present our results on the hold-out test set.
 Since the data is imbalanced, we selected the models based on the best AUC-ROC and AUC-PRC from the validation set. We also evaluate the models based on precision, recall, and brier score~\cite{rufibach2010use}. Deep learning networks are known to be under-calibrated and the last metric measures how well the model is calibrated, i.e., it measures the probabilistic interpretation of risk prediction. In other words, if a model predicts a $0.7$ risk for a patient, brier score measures how well that translates to a $70\%$ chance of the patient developing the complication.
The hyper-parameters for the deep-learning models were selected using a grid search strategy varying batch sizes  $\lbrace 8, 16, 32, 128 \rbrace$, number of layers $\lbrace 1, 2, 3 \rbrace$, and dropout $\lbrace 0.0, 0.1, 0.2 \rbrace$ along with standard initialization and using ADAM as the optimizer of choice.

\subsection{Post-hoc Explainer Models}
% \todotip[PC]{lead}

While some of the classical algorithms considered in Section~\ref{ssub:risk_model} are inherently interpretable with easy access to the features deemed important for the model (such as LR), several of the deep learning models are black-box models. 
To extract feature importances from such models, we used post-hoc explainers which have been found to be favored by clinicians in past studies~\cite{tonekaboni2019clinicians}.
In particular, we used the well accepted SHAP algorithm~\cite{lundberg2018explainable} to find feature importance~\footnote{In this paper, our primary goal is study the importance of contextual explanations and thus we chose SHAP as a well-known SOTA post-hoc explainer. However, we caution the readers about known criticism of SHAP, and in general explainability methods, that are still active area of research without a common consensus}. The algorithm uses game-theoretic principles to identify importance of features by ascertaining the dip in performance of the model with and without access to the feature at the personalized level. 
Such personalized feature importance is key so that our overall risk prediction presentations are more actionable for the clinicians by allowing them to focus on the particular attributes of the patients that are driving their risk.

Typically, clinician time is costly and hard to obtain. Thus to conduct the expert panel sessions and let them focus on some of the most `interesting' patients, we apply Protodash~\cite{gurumoorthy2019efficient} to select a subset of patients. 
Protodash is a post-hoc sample selection method used to obtain a set of prototypical or representative patients from the high risk category that naturally spans the varied set of patient characteristics for the selected sub-group. This also allows the clinician to build trust in using the AI models by inspecting the different patient modalities of the dataset without having to inspect the entire dataset.

\subsection{Extracting Contextual Explanations from Clinical Guidelines}
\label{sub:methods:llm}

\begin{figure*}[!htbp]
    \centering
    \includegraphics[width=\linewidth]{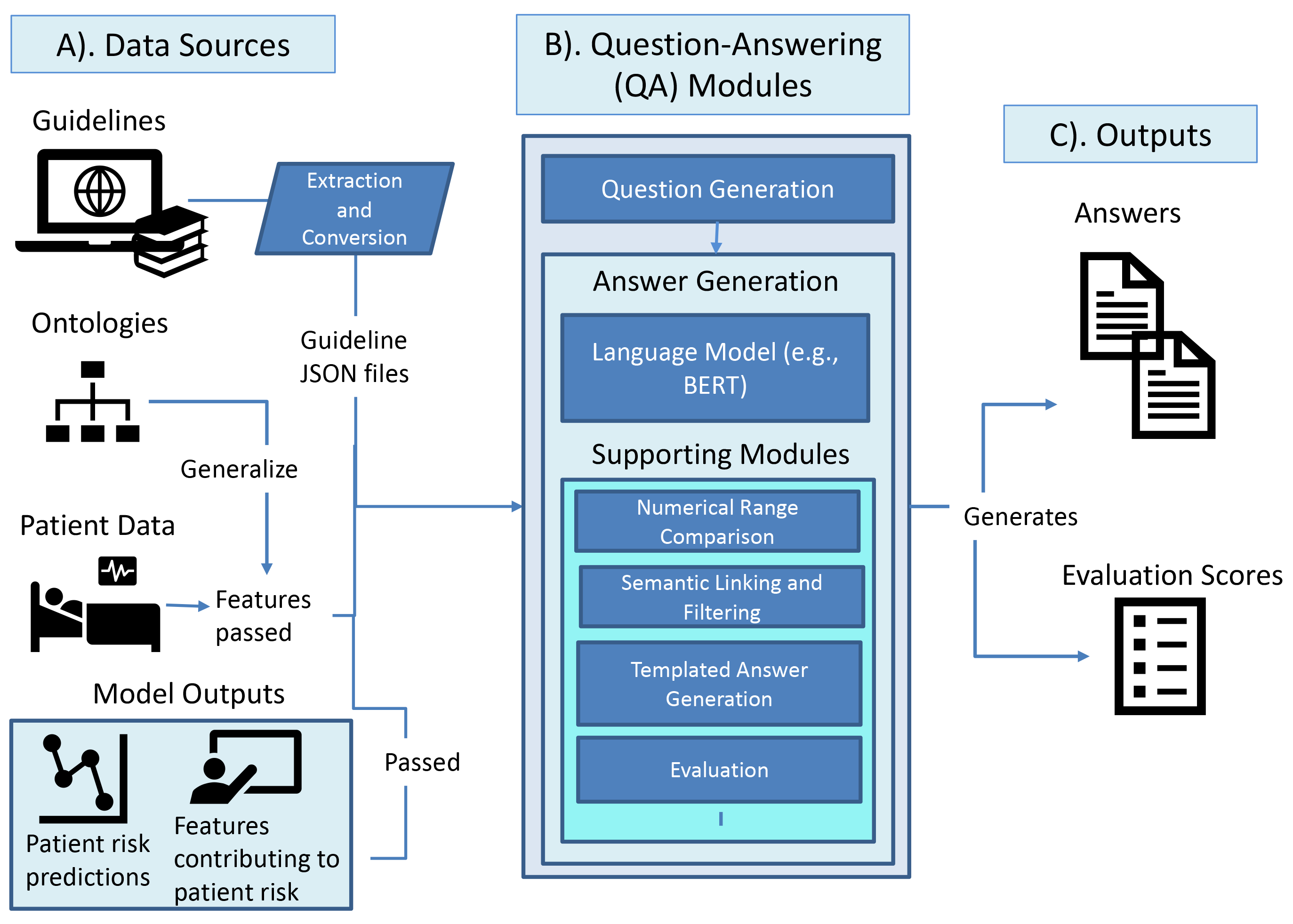}
        \caption{Visualization of different modules within our Question Answering (QA) pipeline including A). information extraction modules, B). QA modules and submodules and C). output modules.}
        %SC not sure if the question bank bit belongs here
        \label{fig:guidelinearchitecture}
\end{figure*}
%OS Add integration and explain how multiple components of multi-method explain risk prediction
%SC reworded to task to remove cyclic nature
 We intend to support a set of possible clinical questions around risk prediction setting for patients. The explanations to these questions can help provide more context around patient predicted risk, features contributing to it and data. Each of these question sets, or question types as seen in Tab. \ref{tab:questiontypes}, can be addressed by multiple sources. %domain knowledge sources. 
 Critically, we set up our problem of extracting context around entities of interest in a risk prediction setting from clinical guidelines to help clinicians make sense of comorbidity risk prediction scores of chronic disease as a question-answering (QA) task.
 %Hence, our QA approach needs to support a diverse set of question types for each patient, which can be answered from multiple knowledge sources, and thus, we have re-purposed some and built submodules that help us address these types of questions. 
Figure~\ref{fig:guidelinearchitecture} shows a detailed overview of the steps involved in this `guideline QA' task. In this section, we give a brief overview of the important steps. For a detailed description please refer to  \ref{appendix:qa_architecture}.

\begin{table*}[!htbp]
  \caption{Question types currently supported by our QA module, we also indicate the data sources used to address questions of the type. } 
  \label{tab:questiontypes}
  \small
  \begin{tabular}{lllc}
    \toprule
    {} & Question Type & \multicolumn{1}{m{3cm}}{Contextualized Entity} & \multicolumn{1}{m{5cm}}{Domain Knowledge Source} \\
    \midrule
    %One-class SVM & & \\
    1 & \multicolumn{1}{m{4cm}}{Patient's {\dm} summary} & \multicolumn{1}{m{3cm}}{Patient} & \multicolumn{1}{m{5cm}}{Patient data}\\
    2 & \multicolumn{1}{m{4cm}}{Patient's risk summary} & \multicolumn{1}{m{3cm}}{Risk Prediction} & \multicolumn{1}{m{5cm}}{Risk Prediction and population data}\\
    3 & \multicolumn{1}{m{4cm}}{Features contributing to patient's {\ckd} risk} & \multicolumn{1}{m{3cm}}{Post-hoc Explanation} & \multicolumn{1}{m{5cm}}{Feature importances and ADA guidelines}\\
    4 & \multicolumn{1}{m{4cm}}{Patient's medication list} & \multicolumn{1}{m{3cm}}{Patient and Risk Prediction} & \multicolumn{1}{m{5cm}}{Patient Data and guidelines}\\
    5 & \multicolumn{1}{m{4cm}}{Patient's lab values} & \multicolumn{1}{m{3cm}}{Patient} & \multicolumn{1}{m{5cm}}{Patient Data and guidelines}\\
  \bottomrule
\end{tabular}
\end{table*}
%Some points
% Blind evaluation - on the model selected of the weak labels (annotated by us)
% If we need to ask Fernando for evaluation, need a schedule; can we pre-generate a list of answers or are we waiting on semantic filtering answers

%\begin{itemize}
\par \noindent  \ul{Information Retrieval from Data Sources:} We support the extraction of context from three domain sources in our QA approach, including patient data, medical ontologies like Clinical Classification Software (CCS) codes~\footnote{\url{https://www.hcup-us.ahrq.gov/toolssoftware/ccs10/ccs10.jsp}} and medical guidelines from ADA Standards of Care 2021 (as introduced in Sec. \ref{sec:background}). We query patient data from Limited Claims Explorys Dataset (LCED) claim records (see Sec. \ref{sec:use-case}) on-demand, either when we need to create questions based on patient parameters or when we need to include these patient values in answers to questions about the patient.  We extract content from the HTML or web version of the `Standards of Medical Care in Diabetes'~\cite{care2021standards} guidelines, published by the American Diabetes Association (ADA)~\footnote{\url{https://care.diabetesjournals.org/content/44/Supplement_1}} using a Python library, BeautifulSoup~\cite{richardson2007beautiful}. We also query patient risk predictions and feature importances using a unique identifier, the patient ID. Some of the extracted contexts are used for generating questions such as patient data, risk predictions and feature importances, and others are used to query against such as the extracted guidelines.
\par \noindent \ul{Question Answering Steps}: Here we describe part B). of our QA architecture (Fig. \ref{fig:guidelinearchitecture}), including the question and answer generation modules and their supporting submodules. In our QA setup we leverage SOTA LLMs and introduce knowledge augmentations to improve their performance on the ADA 2021 medical guidelines. Additionally, we introduce sub-modules to enhance the LLMs' capabilities to address question types 3 - 5 from Tab. \ref{tab:questiontypes}, i.e., diagnosis codes, drugs and clinical indicators, which are run against our extracted guideline content. Below are the submodules in our QA setup:
\par \textit{Question Generation:} The \textit{question generation module} almost always creates templated questions using Python's native support for String Templates,~\footnote{\url{https://docs.python.org/3/library/string.html}} and does so based on patient data, more specifically from the patient's diagnoses codes, lab values, and medication list. We also support the creation of two standard, non-variant questions for each patient, i.e., whose values don't change from patient data, that can help clinicians easily interpret their predicted risk (question type 1) and their {\dm} state (question type 2). Moreover, as can be seen from Tab. \ref{tab:questiontypes}, each of the question types that we support on a per patient basis is populated from different data sources. Hence, we have developed different answering methods for each, including simple lookups and knowledge augmented language model capabilities, including combinations of either a LM + value range comparison or LM + semantic filtering. We provide examples of questions and answers for each question type in \ref{sec:result_settings:appendix}.
\par \textit{Answer Generation:} For questions types 1 and 2 from our supported question types, whose context does not depend on guidelines as shown in Tab. \ref{tab:questiontypes}, we query patient data and feature importances and use a similar templating approach for question generation to populate answer templates with the retrieved query results. For other question types 3 - 5 that are answered by guideline content, we pass them through our LLM and knowledge-augmented LLM setup that we describe next.
\par \textit{Language Models for Generating Answers:} We use a LLM approach in order to find answers to our questions with the unstructured and natural language discussion and recommendation sentences of the ADA 2021 guidelines. We have applied the original Bidirectional Encoder Representations from Transformers (BERT) model or BERT~\cite{devlin2018bert} and other variants of the same retrained on clinical datasets, including SciBERT~\cite{otegi-etal-2020-automatic}, BioBERT~\cite{lee2020biobert}, BioBERT-ASQ~\cite{Yoon2019PretrainedLM} and BioClincalBERT-ADR~\cite{huggingfaceBioClinicalBERT-ADR}. All of the models we utilize are available on the HuggingFace~\cite{wolf2019huggingface} model repository, and we choose BERT models that were made available specifically for clinical question-answering. We built two other submodules to enhance the capabilities of the LLM approach, specifically to address questions with numerical comparisons of question type 5, and to improve the semantic match between the question and the answers returned by a LLM (question type 3 and 4). For details on standard processing steps to include more data types like numerical ranges, refer to Sec. \ref{appendix:qa_architecture} in the appendix.
\par \textit{Augmenting Knowledge to LLM:}  Transformer based LLM approaches like BERT and its variants, work on sequences of words that are often seen together and their surrounding words, but don't leverage the semantics of whether these words are diseases, medications, or biological processes. We found that in the absence of this semantic knowledge, we would often get answers from BERT that don't correlate on a semantic level with the question.  To eliminate such answers, we explored options for a biomedical semantic mapper and zeroed in on the National Library of Medicine (NLLM)'s Metamap tool~\cite{aronson2010overview}. We choose Metamap because of its extensive coverage of biomedical semantic types and its ability to capture entity mentions within the ADA 2021 CPG. Within our pipeline, we have integrated a Python wrapper for Metamap\footnote{PyMetamap: \url{https://github.com/AnthonyMRios/pymetamap}} that can recognize biological entities within the guideline text and their semantic types (e.g., dsyn: disease or syndrome, phsu: pharmacolgic substance, etc. for a complete list of types returned by Metamap see: \footnote{\url{https://lhncbc.nLLM.nih.gov/ii/tools/MetaMap/Docs/SemanticTypes_2018AB.txt}}). Additionally, given this ability to filter based on semantic types, we want to allow additional answers with mentions of related diseases. To provide more broad answers, we use the UMLS Concept Unique Identifier (CUI) codes from the Metamap returned outputs to map to Snomed-CT disease codes~\cite{donnelly2006snomed}. From the mapped Snomed-CT disease codes, we can traverse the Snomed-CT disease tree to identify how many hops apart question and answer disease codes are and if the answer codes are an ancestor of those in the question. We operate on the idea that answers about the parent disease code apply to children nodes. We use the outputs of these knowledge augmentation modules to both pre-filter and post-sort LLM model answers for question types 3 and 4 from Tab. \ref{tab:questiontypes}. We report the accuracies for answers that use these knowledge augmentations in the results section (Sec. \ref{sec:results}). For more details on how we used the Metamap and Snomed codes as knowledge augmentation methods within our QA setup, refer to \ref{appendix:qa_architecture}.
Below in Tab. \ref{tab:appendix:questiontypeexamples} and \ref{tab:appendix:questiontypeexamples2}, we present sample questions and answers for each question type to provide examples of questions and extracted answers supported by our QA approach. We intentionally don't show patient values in these examples to be compliant with HIPAA restrictions.

\begin{table*}[!htbp]
  \caption{Sample questions and answers for each question type supported by our question-answering approach. Answers such as these serve as contextual explanations that provide information to interpret risk predictions better. We don't provide patient values here due to HIPAA restrictions.} 
  \small
  \label{tab:appendix:questiontypeexamples}
  \begin{tabular}{lllc}
    \toprule
   \multicolumn{1}{m{3cm}}{Question Type} & \multicolumn{1}{m{4cm}}{Sample Question} & \multicolumn{1}{m{6cm}}{Answer} \\
    \midrule
    %One-class SVM & & \\
     \multicolumn{1}{m{3cm}}{1. Patient's {\dm} summary} &  \multicolumn{1}{m{4cm}}{What is the patient's A1C value? What are their most frequent diagnoses codes?} & \multicolumn{1}{m{6cm}}{Patient's A1C is A. Their most frequent diagnosis codes are essential hypertension, septicemia, etc.}\\
     \midrule
    \multicolumn{1}{m{3cm}}{2. Patient's risk summary} & \multicolumn{1}{m{4cm}}{How does the predicted risk of the patient compare against the population?} & \multicolumn{1}{m{6cm}}{The predicted risk of chronic kidney disease the patient is X \%. The population averages for the same condition are as follows: For Medicare patients: Y \%  For patients with Charlson Comorbidity Index (CCI) score of 3 : Z \%}\\
    \midrule
    \multicolumn{1}{m{3cm}}{3. Features contributing to patient's {\ckd} risk} & \multicolumn{1}{m{4cm}}{What can be done for Essential Hypertension?} & \multicolumn{1}{m{6cm}}{10.3 For patients with diabetes and hypertension, blood pressure targets should be individualized through a shared decision-making process that addresses cardiovascular risk, potential adverse effects of antihypertensive medications, and patient preferences. C}\\
  \bottomrule
\end{tabular}
\end{table*}

\begin{table*}[!htbp]
  \caption{Sample questions and answers for each question type supported by our question-answering approach. Answers such as these serve as contextual explanations that provide information to interpret risk predictions better. We don't provide patient values here due to HIPAA restrictions.} 
  \label{tab:appendix:questiontypeexamples2}
  \small
  \begin{tabular}{lll}
    \toprule
   \multicolumn{1}{m{3cm}}{Question Type} & \multicolumn{1}{m{4cm}}{Sample Question} & \multicolumn{1}{m{6cm}}{Answer} \\
    \midrule
    \multicolumn{1}{m{3cm}}{4. Patient's lab values} & \multicolumn{1}{m{4cm}}{What should be done for this patient, whose A1C levels are greater than 10 ?} & \multicolumn{1}{m{6cm}}{The early introduction of insulin should be considered if there is evidence of ongoing catabolism (weight loss), if symptoms of hyperglycemia are present, or when A1C levels are greater than 10\% [86 mmol/mol] or blood glucose levels greater than or equal to 300 mg/dL [16.7 mmol/L] are very high.}\\
    \midrule
    \multicolumn{1}{m{3cm}}{5. Patient's medication list} & \multicolumn{1}{m{4cm}}{What do the guidelines state about the GLP-1 RA drug the patient is taking?} & \multicolumn{1}{m{6cm}}{ Meta-analyses of the trials reported to date suggest that GLP-1 receptor agonists and SGLT2 inhibitors reduce risk of atherosclerotic major adverse cardiovascular events to a comparable degree in patients with type 2 diabetes and established ASCVD (185).}\\
  \bottomrule
\end{tabular}
\end{table*}

\subsection{Prototype Dashboard and Expert panel sessions} \label{sec:dashboard}
\begin{figure*}[!htbp]
    \centering
    \includegraphics[width=\linewidth]{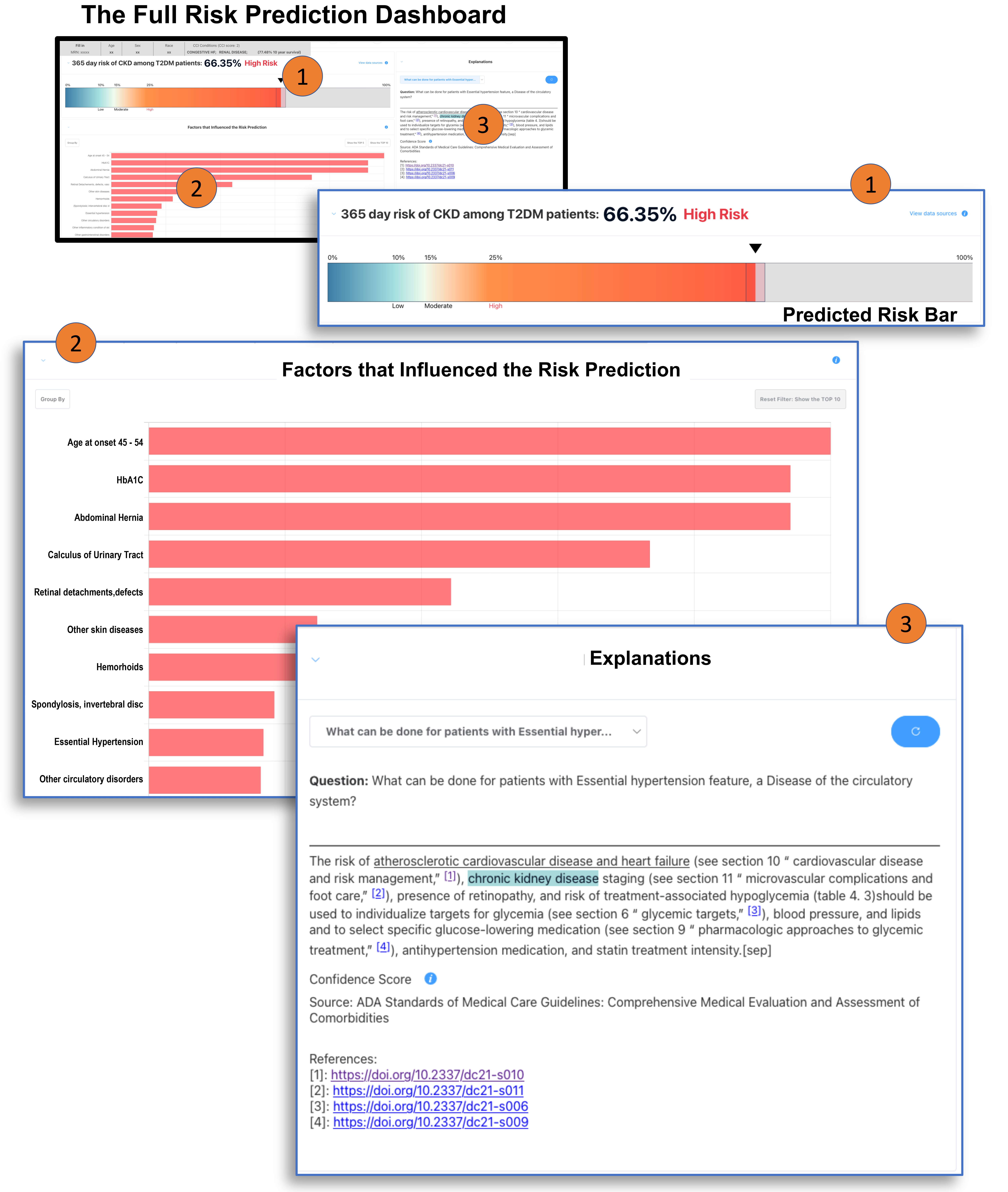}
        \caption{A screenshot of a running prototype of our risk prediction dashboard which includes: 1) the risk prediction score, 2) the features contributing to the predicted risk with the size of their impact on the model results, and 3) a "questions in context pane", in which the user can select and see answers to questions that provide additional contextual information about the patient, the predicted risk calculation, and individual features contributing to the risk.}
        \label{fig:prototypedashboard}
\end{figure*}
% \todotip[SC]{verify these statements and complete it}
%SC reads fine

To present the supported contextual explanations, 
%we support from our multi-method approach, alongside the other patient data relevant to the patient's predicted risk scores, 
we have adapted a question-driven design~\cite{liao2020questioning} for user-interface (UI) development and built a running prototype of a risk prediction dashboard (as seen in Fig.~\ref{fig:prototypedashboard}). The content we show on it is rendered on a per-patient basis chosen from a landing page not shown here. For each patient, we show multiple panes (or UI sections) at a high level, each of which displays content under a particular grouping. These panes include groupings of \textit{patient details, history timeline of claim incidences, risk prediction scores, features contributing to risk, and questions in context}. In Fig. \ref{fig:prototypedashboard}, we highlight the risk prediction, feature importance, and questions in context panes. The explanations pane serve as a section where our contextual explanations, that provide context around our identified entities of interest in the risk prediction setting - patients, their predicted risk, and the features contributing to risk - can be selected and browsed. Additionally, as we have described in Sec. \ref{sec:background}, risk scores can be interpreted better in the context of use, i.e., by enabling connections to patient data, feature importance, and domain knowledge, hence, we had to support interactions between these panes (refer \ref{sec:dashboard:appendix} for the details), which would make it easier for clinicians to establish the connections.

\subsubsection{Expert Panel Sessions using Prototype Dashboard as an Aid}
We used our risk prediction dashboard as an aid during our structured feedback sessions, where we walked clinicians through a live demonstration of our dashboard for a set of prototypical patients (see Tab. \ref{tab:protosummary_appendix}). We conducted sessions individually with four clinicians in our expert panel to understand whether the contextual explanations provided, patient predicted risk, and risk explanations, were helpful for clinical practice. We \textit{explained that this dashboard would be available in addition to the clinician's regular EHR tools and the patient information they provide}, and is meant specifically to provide additional information related to the CKD Risk Prediction. To strike a balance between limited clinician time and the need for diverse feedback, we generated such reports from among $20$ prototypical {\ckd} high-risk patients from our {\dm} cohort, identified by the Protodash algorithm~\cite{gurumoorthy2019efficient}. 
%Each member of the expert panel were shown a maximum of $3$ patient reports and presented a usage scenario. 

During the sessions, we first familiarized each clinician with the different sections of the risk-prediction dashboard. We asked them to imagine that they would be meeting with the patient and had seen the CKD prediction, and stated we wanted to understand what information would be useful to them in understanding the prediction and its impact on their treatment decisions. We then presented the dashboard as it would appear for $3$ randomly selected prototypical patients. We asked the panel members to imagine that they were preparing to treat a patient that was new to them.  We navigated through the dashboard as instructed by the subjects, opening sections or clicking on items as they requested.  We asked the clinicians to speak aloud as they were working with the dashboard.  We also probed the relevance and usefulness of the different sections of the dashboard and the specific content shown in them.  We asked if there was other information they would have liked to have been provided, or questions they would want answered. Sessions were recorded and transcribed, similar to the approach mentioned in ~\cite{knoll2022user}. 

Through these sessions, we wanted to understand the usefulness of our supported patient contextualizations and the features contributing to their risk. Specifically, we showed clinicians the content on different panes of this dashboard and the supported interactions to understand  what features were most important, both from a UI and informational perspective. We report the results of these interactions in Sec. \ref{sec:studyresults}. 
% \todotip[SC]{add a paragraph about the context for this dashboard i.e. Add to -> This is going to be used in complement to patient EHR data, not independently. They evaluated it based on how useful the information was and looked for actionable insights.}
%SC Dan addressed this

\begin{table}
\caption{Summary (generated using Tableone library~\cite{pollard2018tableone}) of  $20$ prototypical patients highlighting the demographic and diagnoses counts. We report the disease diagnoses by their higher-level disease groupings (e.g. for {\dm} the higher-level code is endocrine, nutritional and metabolic disorders). We highlight the conditions that are most prevalent amongst the patients ($> 50\%$).}
% -- CASE WHEN age_at_onset>=19 AND age_at_onset<=44 THEN 1 ELSE 0 END AS AGE_GRP_Y,
% -- CASE WHEN age_at_onset>=45 AND age_at_onset<=54 THEN 1 ELSE 0 END AS AGE_GRP_M,
% -- CASE WHEN age_at_onset>=55 AND age_at_onset<=64 THEN 1 ELSE 0 END AS AGE_GRP_O
\centering
\small
\label{tab:protosummary_appendix}
 \begin{tabular}{lc}
\toprule
                                                                       Feature                  &     Overall counts ($\%$) \\
\midrule
%n &          20 \\
Age at onset 45-54 &    4 (20.0) \\
\textbf{Age at onset} $\geq$ 55 &   \textbf{15 (75.0)} \\
Age at onset $\leq$ 44 &     1 (5.0) \\
SEX - FEMALE &    7 (35.0) \\
\midrule
Mood disorders  &    3 (15.0) \\
Diseases of the blood and blood-forming organs  &    3 (15.0) \\
\textbf{Diseases of the circulatory system} &   \textbf{17 (85.0)} \\
Diseases of the digestive system &    6 (30.0) \\
Diseases of the genitourinary system &    9 (45.0) \\
\begin{tabular}[c]{@{}l@{}} \textbf{Diseases of the musculoskeletal system and} \\ \textbf{ connective tissue} \end{tabular}  &   \textbf{12 (60.0)} \\
Diseases of the nervous system and sense organs  &    9 (45.0) \\
\textbf{Diseases of the respiratory system} &   \textbf{11 (55.0)} \\
Diseases of the skin and subcutaneous tissue &    7 (35.0) \\
\begin{tabular}[c]{@{}l@{}} \textbf{Endocrine; nutritional; and metabolic diseases and} \\ \textbf{ immunity disorders} \end{tabular} &  \textbf{20 (100.0)} \\
Infectious and parasitic diseases &   10 (50.0) \\
Injury and poisoning &    4 (20.0) \\
Mental Illness &    3 (15.0) \\
Neoplasms  &    6 (30.0) \\
\begin{tabular}[c]{@{}l@{}} Symptoms; signs; and ill-defined conditions \\ and factors influencing health status \end{tabular} &   10 (50.0) \\
\bottomrule
\end{tabular}
\end{table}

\section{Results and Evaluation Study}
\label{sec:results}
In this section, we present quantitative results for the guideline question-answering methods (Sec. \ref{sec:guidelinecoverage} and Sec. \ref{sec:guidelineqaresults}). As a qualitative analysis, we also discuss themes and subthemes that we found from an analysis of discussions from our expert panel sessions (Sec. \ref{sec:studyresults}). For our risk prediction model we choose MLP, and derive important features for the prediction using the SHAP model. 
%\todotip[PC]{See if MLP is mentioned earlier.}
The distribution of important features identified by SHAP can be seen in Tab. \ref{tab:protosummary_appendix}. Results for these models can be browsed via the appendix \ref{sec:result_settings:appendix}.

\subsection{Data coverage and support} \label{sec:guidelinecoverage}
%\subsection{Coverage of the Standards of Care in Diabetes 2021 Guidelines} \label{sec:guidelinecoverage}
% \todotip[SC]{start with this.. complete the following line}
\begin{figure*}[!htbp]
    \centering
    \includegraphics[width=\linewidth]{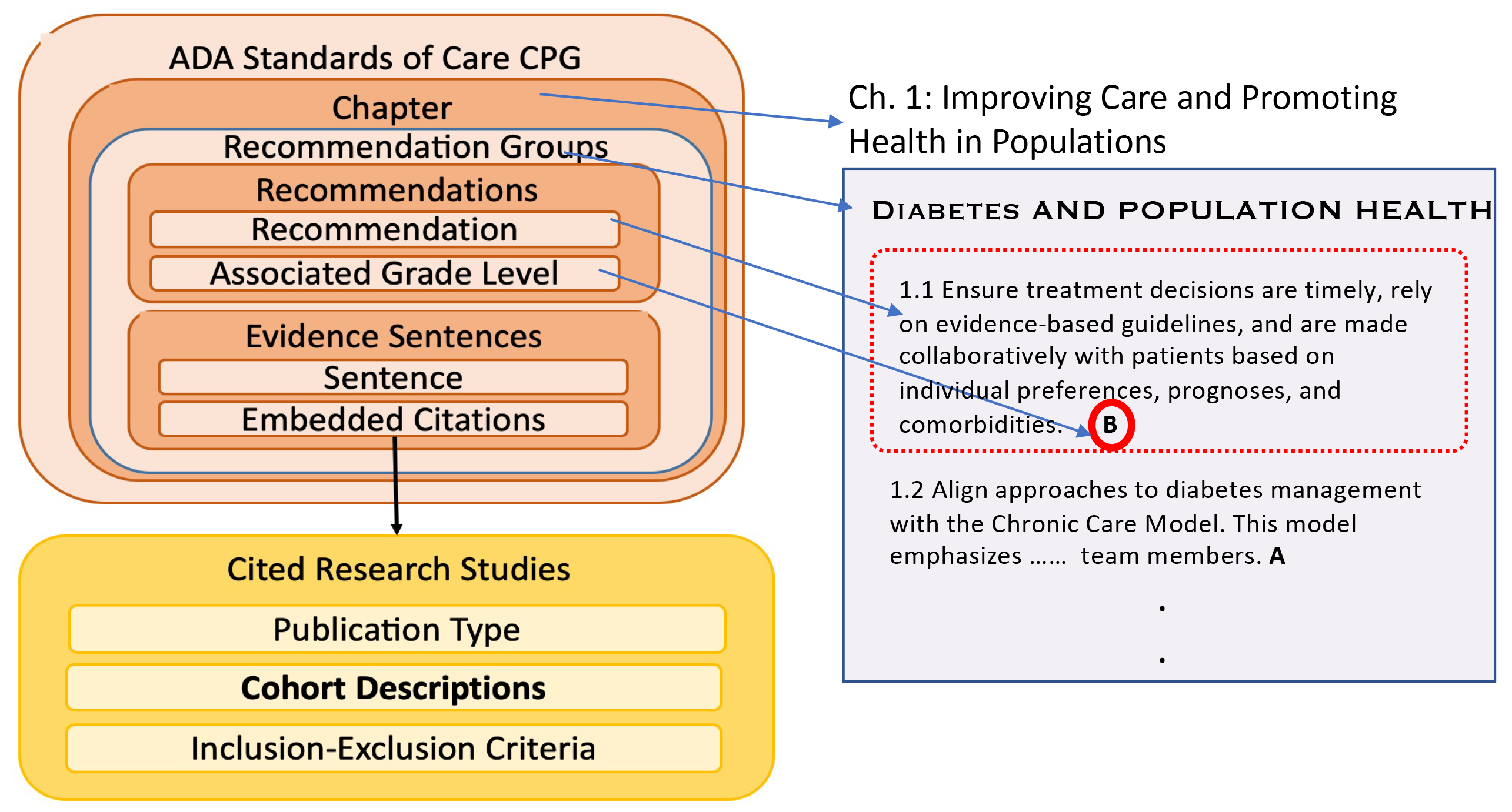}
        \caption{Overview of the evidence structure in the ADA Standards of Medical Care - Diabetes Guidelines 2021.}
        \label{fig:guideline-structure}
\end{figure*}

One of the aims of this manuscript is to extract contextual explanations from medical guidelines. To explore the feasibility of this task we first analyzed the coverage of the ADA 2021 guidelines used to extract contextual explanations of predictions in {\ckd} comorbodity, {\dm} risk prediction setup. 
% \todotip[SC]{complete this line and link sections including table 1, and mentions of risk prediction model and say we focus on how feasible to extract (not support) contextual explanations from guidelines.}
%SC done
%SC for semantic types covered - need to run Metamap on all sentences and extract semantic types per sentence
%SC Avg. tokens per sentence - run NLTK against the JSON dump 
%SC question should we fix open bugs for extraction?

We extracted the recommendations and discussion sentences across the $16$ chapters of the current ADA 2021 CPGs. These recommendation and discussion sentences are expressed in natural language (See Fig. \ref{fig:guideline-structure}). Table~\ref{tab:coverageguidelines} shows a high-level overview of the coverage statistics. Specifically, the extracted sentence corpus can hence be analyzed for the total number of tokens, average token length per sentence, and their composition of Metamap semantic types, to understand the coverage of the guideline text in terms of volume and semantic diversity. For tokens, we report words recognized by both BERT's tokenizer model to be consistent with our QA approach.
%SC commenting for now
% and Python's NLTK library across different parts of speech tags.
~\cite{hematialam2021identifying} report similar statistics for three other CPGs, neither of which are Diabetes focused, but it can be seen that the total number of tokens and sentences in the ADA 2021 CPG are more than the three guidelines reported in this paper. Hence, pointing to the comprehensiveness of our approach. Also, note that some of the recommendations were not captured by our guideline extraction script, and hence our statistics might be lesser than the actual count.~\footnote{In future, we aim to expand our coverage and update our methods to better capture these recommendation groups}.

\begin{table}[!htbp]
  \caption{Coverage statistics from extracted content from the ADA Standards of Care - Diabetes Guidelines 2021. We report these statistics on the recommendations and discussion sentences we extracted across chapters.} 
  \label{tab:coverageguidelines}
  \centering
  \small
  \begin{tabular}{lc}
    \toprule
    Field & Count \\
    \midrule
    %One-class SVM & & \\
    Chapters & 16 \\
    No. of sentences & 2379\\
   Tokens from BERT & 118350 \\
   %SC commenting for now
%   Tokens from NLTK & \\
   Avg. BERT tokens per sentence &  49\\
    Metamap Semantic types covered & 116 / 126\\
  \bottomrule
\end{tabular}
\end{table}

\begin{figure*}[!htbp]
    \centering
    \includegraphics[width=\linewidth]{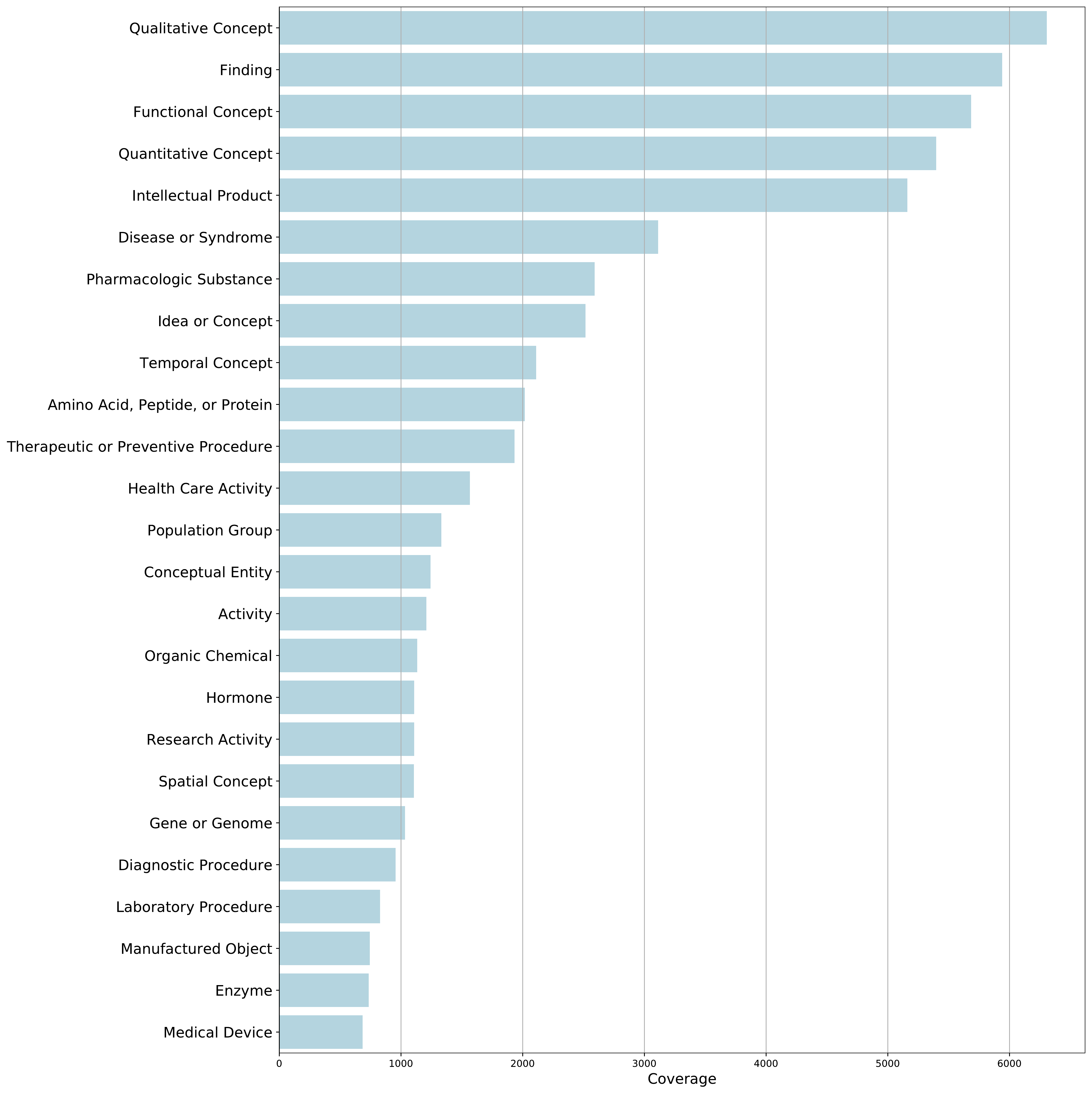}
    \vspace{-2em}
        \caption{Frequency distribution for $25$ / $116$ of the top semantic types that were found in the extracted guideline text.}
        \label{fig:semantic_type_dist}
\end{figure*}

The many semantic types (see Fig. \ref{fig:semantic_type_dist} for 25 most populous semantic types) covered by the ADA 2021 CPG reaffirms that guidelines are a comprehensive source of evidence-based information in the clinical domain~\cite{murad2017clinical}. Further, in Sec. \ref{sec:discussion}, we also discuss how well the ADA 2021 CPG alone can support the themes we analyzed from conversations with clinicians during our structured feedback sessions and hence discuss the CPG's ability to serve as a source of context in our risk prediction setting.

\subsection{Quantitative Evaluation of Guideline QA} \label{sec:guidelineqaresults}
One of our key aim is to study whether SOTA LLM methods can be used to extract high quality contextual explanations. 
Thus, while we can address different question types in our QA approach, as seen in Tab. \ref{tab:questiontypes}, we only evaluate those question types that are addressed by ML methods, i.e., through our knowledge and knowledge augmented LLM modules (as described in Sec. \ref{sec:methods} and seen in Fig. \ref{fig:guidelinearchitecture}). We evaluate feature importance questions of types 3 (of diagnostic importance), type 4 (of treatment importance), and  type 5 (asking about clinical indicators and important for both diagnostic and management purposes). Additionally, we evaluate answers to questions of types 3 and 4 
%SC commenting for now and 5 
differently from question type 5 since questions of types 3 and 4 are served by the knowledge augmented LLM modules and questions of type 5 are addressed by the LLM with the numerical range comparison module. For answers to questions of type 3 and 4, we report standard and frequently used NLP QA metrics including mean average precision (MAP)~\cite{teufel2007overview}, F1-score their contributors of precision and recall and BLEU scores. For answers to questions of type 5, we report the number of times the combination of the LLM and numerical range comparison module could correctly predict whether the answer outputted was in / out range for the numerical value being asked about in the question. 
% \todotip[SC]{remove this. added this line to expert panel, here say something like:  - verify this}
%SC Done
% %We report all results, whether for question types 3 %SC commenting for now and 5 or 4, on the questions generated for the 20 prototypical {\ckd} high-risk patients from our {\dm} cohort, identified by the Protodash algorithm~\cite{gurumoorthy2019efficient}. 

\subsubsection{Evaluation Results of Feature Importance Questions of Diagnostic Importance} \label{sec:resultsquestions}
We first evaluated the quality of extracted contextual explanations for feature importance questions, type 3. As outlined in Sec.{~\ref{sub:methods:llm}}, our aim was to evaluate the readiness of SOTA LLM models for this task.
Here we report the results for $71$ questions covering relevant feature importances for patients' risk predictions and these questions cover $14$ CCS LVL 1 diagnosis code types. As mentioned previously, we report the performance in terms of a number of standard metrics. However,  given our information extraction setting, among these we are especially interested in the precision metrics. These metrics measure how many documents, among the ones retrieved, were relevant. Furthermore, `MAP' and `precision@k' measures the same while considering the order of retrieval. This aligns closely with how clinicians evaluate presented informations where the presented answers are expected to accurate in order of most acceptable to least.  
%To evaluate answers to questions of feature importance questions, question type 3, we tried many settings with different language augmentations from Metamap and Snomed. We used the outputs of the language augmentation modules that included the number of diseases matched between question and answers, number of ancestors of questions in the answer and the number of hops between question and answer disease codes as features to either filter the answer set before passing to BERT, pre-filtering, or sort the answer set from BERT based on these feature values, post-sorting. 
% Further, we have replicated the same set of experiments with a BioBERT model to see if there is an improvement in using a domain-based LLM, and we see an X gain in the best setting. 
It is to be noted, that we evaluated the predicted results to disease feature importance questions against candidate answers by manually inspecting the ADA 2021 CPG. The annotations were done by an author and some of these annotations were verified by a clinical expert on the team who is also a co-author in this paper. We report results on the expert validated subset in \ref{sec:result_settings:appendix}. Table~\ref{tab:nativemodelresults} reports results in comparison to the entire annotated dataset of $85$ feature questions and $654$ candidate answers for the out-of-the shelf LLM models under consideration. The results show that in terms of `MAP' as well as `precision at k' metrics, vanilla BERT outperforms the other LLM models. SciBERT is a close second with an improved recall and F1 score. We analyze the importance of these results further in Sec.~\ref{sec:discussion}.

Additionally, we also evaluated the results by augmenting the base LLM models with different strategies. 
Tab.~\ref{tab:kmdiseaseresults} reports the result scores for answers to question type 3 using the best knowledge augmentation strategy (across the five different settings) for each LLM models. Overall, we can see a significant improvement in `MAP' and `precision at 5' for knowledge augmented BERT model (BERT-KA) over BERT, the best performing base language model. In terms of other metrics, knowledge augmented SciBERT-KA shows a consistent improvement for all metrics while being best/second-best overall.

% \todotip[SC]{Need to fill in these numbers}
%SC done
While these evaluation numbers are reported from a small evaluation set of $71$ questions and $654$ candidate actual answers, we consider this as a somewhat comprehensive evaluation due to the diversity of diseases covered ($1844$ diseases) and in total semantic types covered within the answers ($116$ semantic types, see Tab. \ref{tab:coverageguidelines}), both in the candidate and predicted sets. Overall, from these results, we see that  knowledge augmentation can improve the base language model performance. 
%settings, and that there is value in either filtering the answers to be passed to a LLM like BERT or sorting the answers from BERT, via aids from known domain knowledge sources that data sources like guidelines are expected to adhere to. 
% We also report BLEU scores for these settings in addition to F1 metrics to also indicate how close our answers are to human judgment~\cite{chen2019evaluating}. 

\begin{table}[!htbp]
\caption{Performance of Guideline QA on different language model approaches reported at mean average precision (MAP), F1 and recall at top-10 answers and precision at top-1 and top-5 for $71$ disease feature importance questions. The models are sorted by MAP values, to indicate an ordering of the best models. }

\label{tab:nativemodelresults}
\small
\centering
\begin{tabular}{lrrrrrr}
\toprule
{} & {bleu} & {P@1} & {P@5} & {map} & {f1} & {recall} \\
{model} & {} & {} & {} & {} & {} & {} \\
\midrule
BERT & 0.117 & \color{green} 0.468 & \color{green} 0.382 & \color{green} 0.390 & 0.213 & 0.241 \\
BioBERT & 0.116 & 0.431 & 0.339 & 0.346 & 0.200 & 0.238 \\
BioBERT-BioASQ & \color{blue} 0.132 & 0.383 & 0.329 & 0.332 & \color{blue} 0.217 & \color{blue} 0.281 \\
BioClinicalBERT-ADR & 0.125 & 0.368 & 0.317 & 0.316 & 0.205 & 0.259 \\
SciBERT & \color{green} 0.165 & \color{blue} 0.461 & \color{blue} 0.349 & \color{blue} 0.364 & \color{green} 0.261 & \color{green} 0.354 \\
\bottomrule
\end{tabular}

\end{table}

\begin{table}[!htbp]
\caption{Performance of Guideline QA of knowledge augmented language models reported at mean average precision (MAP), F1 and recall at top-10 answers and precision at top-1 and top-5 for $71$ disease feature importance questions. Here we show the best knowledge augmentation approach per model to indicate highest gains over baseline performance for the native language model approaches we tried. Best and second-best values for each column is highlighted in \textcolor{green}{Green} and \textcolor{blue}{Blue} color, respectively. Language model (e.g. BERT) suffixed with KA represents the corresponding knowledge augmented model (e.g. BERT-KA).}
\label{tab:kmdiseaseresults}
\small
\centering
\begin{tabular}{lrrrrrr}
\toprule
{} & {bleu} & {P@1} & {P@5} & {map} & {f1} & {recall} \\
{model} & {} & {} & {} & {} & {} & {} \\
\midrule
BERT-KA & 0.075 & \color{blue} 0.467 & \color{green} 0.419 & \color{green} 0.438 & 0.169 & 0.186 \\
BioBERT-KA & 0.127 & 0.434 & 0.348 & 0.353 & 0.215 & 0.254 \\
BioBERT-BioASQ-KA & \color{blue} 0.141 & 0.458 & \color{blue} 0.362 & 0.369 & \color{blue} 0.237 & \color{blue} 0.280 \\
BioClinicalBERT-ADR-KA & 0.121 & 0.406 & 0.321 & 0.330 & 0.202 & 0.242 \\
SciBERT-KA & \color{green} 0.192 & \color{green} 0.473 & 0.341 & \color{blue} 0.375 & \color{green} 0.291 & \color{green} 0.405 \\
\bottomrule
\end{tabular}

\end{table}

\subsubsection{Evaluation Results for Drug Questions}
%\todotip[SC]{Replication study}
To demonstrate the flexibility of the LLM setup on various settings that match the coverage of the T2D guideline data (see Fig. \ref{fig:semantic_type_dist}) as well as to test the generalizability of our results. We also report results for $6$ anti-diabetic drug questions of question type 4. 
%  and clinical indicator questions of type 4\todotip[SC]{do we really have clinical indicators here?}
Table~\ref{tab:drugresults} show the results for out-of-the shelf language models as well as the knowledge augmented language models for the metrics of interest.
Overall, we once again found the knowledge-augmented language models to be the best performing ones. In terms of `MAP' and `precision at 5', BERT-KA comes out as the best performing model, where as SCIBERT-KA comes out as either the best or the second-best model for all metrics. It is to be noted that the overall results are significantly better than the ones for disease questions. One possible reason behind this effect may be related to the fact that drugs are referred directly in guidelines and thus QA models are able to pick these sentences with greater efficacy.

\begin{table}[!htbp]
\caption{Performance of Guideline QA with different knowledge augmentations of language model approaches reported at mean average precision (MAP), F1 and recall at top-10 answers and precision at top-1 and top-5 for $6$ anti-diabetic drug feature questions. Best and second-best values for each column is highlighted in \textcolor{green}{Green} and \textcolor{blue}{Blue} color, respectively. Language model (e.g. BERT) suffixed with KA represents the corresponding knowledge augmented model (e.g. BERT-KA).}
\label{tab:drugresults}
\small
\centering
\small
\begin{tabular}{lrrrrrr}
\toprule
{} & {bleu} & {P@1} & {P@5} & {map} & {f1} & {recall} \\
{model} & {} & {} & {} & {} & {} & {} \\
\midrule
BERT & 0.100 & 0.910 & 0.751 & 0.757 & 0.254 & 0.206 \\
BioBERT & 0.100 & 0.726 & 0.643 & 0.635 & 0.231 & 0.192 \\
BioBERT-BioASQ & 0.081 & 0.708 & 0.694 & 0.704 & 0.222 & 0.162 \\
BioClinicalBERT-ADR & 0.075 & 0.593 & 0.614 & 0.597 & 0.192 & 0.146 \\
SciBERT & \color{blue} 0.121 & \color{green} 0.947 & 0.757 & 0.772 & \color{blue} 0.281 & \color{blue} 0.228 \\
\midrule
BERT-KA & 0.099 & 0.900 & \color{green} 0.863 & \color{green} 0.821 & \color{blue} 0.281 & 0.213 \\
BioBERT-KA & 0.083 & 0.802 & 0.704 & 0.720 & 0.234 & 0.170 \\
BioBERT-BioASQ-KA & 0.117 & 0.711 & 0.725 & 0.716 & 0.272 & 0.221 \\
BioClinicalBERT-ADR-KA & 0.085 & 0.598 & 0.595 & 0.587 & 0.199 & 0.152 \\
SciBERT-KA & \color{green} 0.128 & \color{blue} 0.912 & \color{blue} 0.823 & \color{blue} 0.794 & \color{green} 0.298 & \color{green} 0.232 \\
\bottomrule
\end{tabular}

\end{table}

\subsubsection{Evaluation Results of Clinical Indicator Questions}
\begin{table}[!htbp]
  \caption{Results of Guideline QA with rule augmentation of language model approaches for numerical comparisons reported for $9$ questions across the $20$ prototypical patients identified from our predicted high-risk chronic kidney disease cohort. The split of question variations is equal across the different numerical range comparison operators of lesser than, equal to and greater than.} 
  \label{tab:resultsguidelines_numerical}
  \centering
  \small
  \begin{tabular}{lllllll}
    \toprule
    Comparison & Accuracy & TP & TN & FP & FN & Total  \\
    \midrule
     Overall & 0.78 & 7 & 7 & 3 & 0 & 18 \\
     %FP = 1
     Lesser Than & 0.84 & 2 & 3 & 1 & 0 & 6  \\
     %FP = 1
    Equal To & 0.67 & 1 & 3 & 2 & 0 & 6  \\
     %FP = 0 and FN = 0
     Greater Than & 100 & 4 & 2 & 0 & 0 & 6  \\
  \bottomrule
  \multicolumn{7}{m{10cm}}{TP - True Positives, TN - True Negatives, FP - False Positives, FN - False Negatives. Accuracy computed as accuracy = (TP + TN)/ Total} \\
\end{tabular}
\end{table}

In Tab. \ref{tab:resultsguidelines_numerical}, we report the accuracy statistics for the performance of our rule augmentation / numerical range comparison module of our QA approach. We show granular breakdowns based on different numerical operators, greater than, lesser than, and equal to, to indicate which settings are being picked up the best by the rule augmentation and which others are harder. In this evaluation, we manually went through the outputted answers to ensure that they were within range of the numerical values in questions. The reason for this annotation approach is that the guidelines have few sentences for actions to be taken on clinical indicators.
% \todotip[SC]{What is the most performant setting.}
Hence, there is not much diversity in the answers that a LLM like BERT can output before passing the answer to our range comparison module for validation. These accuracies highlight the value in combining syntactic parsing output with a LLM to improve its capabilities. 

Overall,  these quantitative evaluations points to the potential in applying scalable, augmented LLM-based approaches to extract content from authoritative guideline literature that can then be used to provide context to interpret model predictions, such as in our setting, risk prediction scores and their model explanations.

\subsection{Qualitative Evaluation with Clinicians} \label{sec:studyresults}
% \todotip[DG]{Lead}
% \todotip[Dan]{Move some of these to methods.}

% We conducted sessions individually with four clinicians in our expert panel to understand whether the explanations we support that provide contextual explanations around the patient, their predicted risk, and the explanations of their risk, were helpful in clinical practice. We started by familiarizing each clinician with the different sections of the risk-prediction dashboard.  We then presented the dashboard as it would appear for a series of different simulated patients,  loosely based on actual patients with certain details modified or obscured. We asked the panel members to imagine that they were preparing to treat this patient who was new to them.  We navigated through the dashboard as instructed by the subjects, opening sections or clicking on items as they requested.  We asked the clinicians to speak aloud as they were working with the dashboard.  We also probed about the relevance and usefulness of the different sections of the dashboard and the specific content shown in them.  We asked if there was other information they would have liked to have had, or questions they would want answered. Sessions were recorded and transcribed. 

% \todotip[Dan]{Keep in results.}
We conducted a thematic analyses on the responses and feedback received during the expert panel sessions (Sec. \ref{sec:dashboard}), as follows.  Three independent researchers, who are coauthors on this paper, reviewed the transcripts, flagged significant utterances, and characterized these utterances in terms of the major points and themes they expressed.  The researchers then reviewed their sets of identified themes and utterances together, and grouped and combined them into a single agreed-upon set of overarching themes.   We report this combined list of themes below (Tab. \ref{tab:thematicanalysis1}, \ref{tab:thematicanalysis2}, \ref{tab:thematicanalysis3}, \ref{tab:thematicanalysis4}). These themes reflect areas that clinicians prioritize and where the support of explanation-driven, AI risk prediction tools would be appreciated. 

As seen in Tab. \ref{tab:thematicanalysis1}, \ref{tab:thematicanalysis2}, \ref{tab:thematicanalysis3} and \ref{tab:thematicanalysis4}, we have grouped the discussions from the expert panel sessions into \textit{four high-level themes} spanning different areas where clinicians would benefit the most in a chronic disease, comorbidity risk-prediction setting such as ours. The high-level themes we found include `Theme 1: Clinical Value of Explanations and Contextualizations', `Theme 2: Highlighting Actionability', `Theme 3: Connections to Patient Data' and `Theme 4: Connections to External Knowledge'. We were further able to create sub-themes for more granular topics that came up during the discussions under each of these themes, bringing the theme and sub-theme total to {\textit{four high-level themes and twelve sub-themes}}. 

\begin{table}[!htbp]
\begin{center}
\caption{\textit{Clinical Value of Explanations and Contextualizations} - $1^{\text{st}}$ theme that emerged during our expert panel interviews with clinicians where we walked through the risk prediction dashboard and the contextual explanations that we support. We attach a description for each sub-theme that we found and we also provide examples in quotes.}
\label{tab:thematicanalysis1}
\scriptsize
\begin{tabular}{|l|l|}
      %  \multicolumn{2}{c}{\textbf{}}\\ \hline
      \toprule
      \multicolumn{1}{m{4cm}}{Sub-theme} & \multicolumn{1}{m{4cm}}{Description} \\
      \midrule
      \multicolumn{1}{m{4cm}}{Value of Contextual Information around {\ckd} risk} & \multicolumn{1}{m{10cm}}{All clinicians saw value in connecting the {\dm} patient's {\ckd} risks to \textit{data on their other conditions}, and to, \textit{relevant recommendations from the {\dm} guidelines}. For example, some clinicians reasoned about ``how the patient's {\ckd} risk changes their dosage / treatment'' and some others were interested about ``connections to other conditions that patient has.''} \\ \midrule
      \multicolumn{1}{m{4cm}}{Value of Contextual Information around Individual Features} & \multicolumn{1}{m{10cm}}{Clinicians found that information from {\dm} guidelines and cited literature relevant to factors that contributed to the system's predicted {\ckd} risk were helpful to understand how the \textit{factors could be related to {\ckd} or {\dm}}, and how they might interact with \textit{other factors shown}.  For example, ``how does a skull fracture elevate CKD risk?''  This was particularly valuable when not previously known by the clinician, for example:  ``it is surprising, and I have learned something about celiac disease and abdominal pain connection'' in patients with diabetes.}  \\ 
      \midrule
       \multicolumn{1}{m{4cm}}{Value of Contextual Information around patient's {\dm}} & \multicolumn{1}{m{10cm}}{Besides the patient's {\ckd} risk and its implications, the clinicians were interested to know about the \textit{patient's {\dm} state, their comorbid conditions and other parameters in relation to their {\dm} diagnosis}. For example, the clinicians wanted to know ``how long has the patient had their {\dm}'' or ``what is their A1C progression?''} \\ \bottomrule
\end{tabular}
\end{center}
\end{table}

\begin{table}[!htbp]
\begin{center}
\caption{\textit{Highlighting Actionability} - $2^{\text{nd}}$ theme that emerged during our expert panel interviews with clinicians, where we walked through the risk prediction dashboard and the contextual explanations that we support. We attach a description for each sub-theme that we found and we also provide examples in quotes.}
\label{tab:thematicanalysis2}
\scriptsize
\begin{tabular}{|l|l|}
      \toprule
      \multicolumn{1}{m{4cm}}{Sub-theme} & \multicolumn{1}{m{4cm}}{Description} \\
      \midrule
    % \begin{tabular}[c]{@{}l@{}} Better Presentation \end{tabular} & \begin{tabular}[c]{@{}l@{}} The clinicians provided some feedback to \\ improve the content presentation on the dashboard, such as: \\ ``make the references hyperlinks in the answers'',  \\ ``utilize different colors for positive and \\ negative contributors to risk'', etc. \end{tabular} \\ \hline
     \multicolumn{1}{m{4cm}}{Highlight Actionable and Modifiable Factors} & \multicolumn{1}{m{10cm}}{Most of the clinicians were interested in highlighting patient risk factors that could be controlled or acted upon, vs. those (such as age) that could not be influenced: ``what factors can be changed?''} \\ \midrule
           \multicolumn{1}{m{4cm}}{Highlight the Impact of CKD risk prediction on Treatment Decisions for Diabetes and other conditions} & \multicolumn{1}{m{10cm}}{When shown information from treatment guidelines, the clinicians wanted to understand how they \textit{reflect} \textit{the patient's CKD risk and T2D diagnosis}:``are any of these proposed medications contraindicated?''} \\ \midrule
      \multicolumn{1}{m{4cm}}{Suggest Specific Actions to Reduce CKD risk } & \multicolumn{1}{m{10cm}}{Clinicians wanted to understand ways to reduce the CKD risk including ways of addressing risk factors and changes to medications: ``do any of the patient's current medications increase risk of renal toxicity?'} \\ \bottomrule
\end{tabular}
\end{center}
\end{table}

%Solution taken from this post: https://stackoverflow.com/questions/790932/how-to-wrap-text-in-latex-tables
\begin{table}[!htbp]
\begin{center}
\caption{\textit{Connection to Patient Data} - $3^{\text{rd}}$ theme that emerged during our expert panel interviews with clinicians, where we walked through the risk prediction dashboard and the contextual explanations we support. We attach a description for each sub-theme that we found and we also provide examples in quotes.}
\label{tab:thematicanalysis3}

\scriptsize
\begin{tabular}{|l|l|}
      \toprule
      \multicolumn{1}{m{4cm}}{Sub-theme} & \multicolumn{1}{m{4cm}}{Description} \\
      \midrule
   \multicolumn{1}{m{4cm}}{Connections to Patient's Clinical Indicators} & \multicolumn{1}{m{10cm}}{The clinicians indicated that they want to see clinical indicators for diagnoses ( if available ), when interpreting the \textit{factors} \textit{that led to the risk} or the \textit{patient's CKD risk score}: ``Given the patient has essential hypertension, what was their lab systolic blood pressure reading,'' or ``The patient's eGFR value will be important to show for CKD,''  or  ``what do the guidelines say about this patient's systolic and diastolic blood pressure readings.''} \\ \midrule
    \multicolumn{1}{m{4cm}}{Need for Information on Related Diagnoses} & \multicolumn{1}{m{10cm}}{When shown factors which were diagnosis codes that influenced the risk prediction, clinicians wanted to \textit{see what other diagnoses} \textit{ that the patients had, that might align or contribute}: ``show COVID-19 answers for lower respiratory disorders?'' or ``what episodes of abdominal pain did the patient have?''}\\ \midrule
    \multicolumn{1}{m{4cm}}{Connections to Patient's History} &  \multicolumn{1}{m{10cm}}{When shown certain factors, the clinicians wanted to know the when the patient had the diagnosis,  if the condition was a current one, and about changes over time. for example, ``when did the patient have a genitourinary diagnosis?'', or``what does their eGFR progression look like?' } \\ \bottomrule
 \end{tabular}
\end{center}
\end{table}   
    
\begin{table}[!htbp]
\begin{center}
\caption{\textit{Connection to External Medical Domain Knowledge} - $4^{\text{th}}$ theme that emerged during our expert panel interviews with clinicians, where we walked through the risk prediction dashboard and the contextualization we support. We attach a description for each sub-theme that we found and we also provide examples in quotes.}
\label{tab:thematicanalysis4}
\scriptsize
\begin{tabular}{|l|l|}
      \toprule
      \multicolumn{1}{m{4cm}}{Sub-theme} & \multicolumn{1}{m{4cm}}{Description} \\
      \midrule
     \multicolumn{1}{m{4cm}}{Links to Medication Databases} &  \multicolumn{1}{m{10cm}}{When deciding upon \textit{treatment suggestions} for patients given the \textit{knowledge of their {\ckd} risk and {\dm} diagnosis},the clinicians wanted to understand how their medications interact : ``what drugs they are on currently have a bad renal impact?''  or ``how does their current anti-diabetic drug interact with a {\ckd} drug?''} \\ \midrule
     \multicolumn{1}{m{4cm}}{Links to Published Articles} & \multicolumn{1}{m{10cm}}{When connections between the \textit{{\ckd} risk prediction} and \textit{the factors contributing to the risk} were unclear, clinicians mentioned they would look for published references: ``what is the connection between {\ckd} and respiratory disorders?'' or ''how does celiac disease mention from the guideline answer, affect {\ckd}?''} \\ \midrule
      \multicolumn{1}{m{4cm}}{Support familiar categorizations} & \multicolumn{1}{m{10cm}}{Some clinicians were looking for more \textit{provenance around the categorization schemes} we were utilizing to show higher-level physiological pathways for diagnoses codes, and were also hoping for connections to familiar schemes like ``ICD-10'': ``how are hemorrhoids linked to the circulatory system?''  } \\ \bottomrule
\end{tabular}
\end{center}
\end{table}

More specifically, under `Theme 1: Clinical Value of Explanations and Contexts', we group instances where clinicians could make sense of the risk predictions and post-hoc explanations by the additional context provided, or instances where clinicians would appreciate more context. Within `Theme 2: Highlight Actionability,' we discuss instances where clinicians mentioned a need to depict actionable features and indicate actions for them concerning the patient's {\dm} diagnoses or their elevated {\ckd} risk. Under `Theme 3: Connections to Patient Data,' we cover instances where clinicians looked for connections to patient history or their lab results while reasoning about the patient case. Finally, under `Theme 4: Connections to External Knowledge,' we describe instances where clinicians mentioned a need to make connections to the latest literature, other medication databases, or other clinical schemes they utilize. In addition, these themes have an order among them, to address the clinical value of explanations from Theme 1 and the actionability aspects from Theme 2, the content requests from Themes 3 and 4 would be contributing features, which are the connections to patient data from Theme 3 and the links to external knowledge from Theme 4. We present deeper breakdowns in the form of sub-themes and descriptions for each of these four themes in Tab. \ref{tab:thematicanalysis1}, \ref{tab:thematicanalysis2}, \ref{tab:thematicanalysis3} and \ref{tab:thematicanalysis4}.

%Can move this next paragraph to discussion if need be.
As the expert panel sessions were conducted mid-way through our current implementation, some of these themes served as a means for further refinements, such as features to support a `need for more related diagnoses,'  and `connections to patient's clinical indicators,' 
% and `links to published articles.' 
To address these themes, we refined different modules of our pipeline, including the user interface, the question-answering, and post-explanation modules. 
% In Sec. \ref{sec:guidelineqaresults}, we show to what extent the diabetes guidelines alone can support the requests in these themes and sub-themes. 
In future, to address the requests under `Theme 4: Connections to External Knowledge,' we plan to support connections to external resources, which clinicians may find valuable.

In summary, these themes and sub-themes from the expert panel sessions validate our hypothesis about the need for additional clinical context to situate risk predictions and span requests for better connections and presentations of domain knowledge that clinicians are familiar with in these settings.

\section{Discussion} \label{sec:discussion}
%SC discussion is an interpretation of results
% Claims \todotip[SC]{Can ignore and merge with below}
% \begin{itemize}
% \item Context helps and is needed to explain explanations
% \item guidelines contain the most updated systematic evidence in the field 
% \end{itemize}
%Backup discussion with numbers
%SC coverage of guidelines might move here
% \todotip[SC]{mention overlap between themes and guidelines}

%SC extra summary content 
% \textbf{What have we designed?} To support the need for user-centered explainability and make ML predictions and model explanations more useful and clinically relevant~\cite{tonekaboni2019clinicians}, we have designed a multi-method approach to support contextual explanations from domain knowledge sources including medical guidelines, biomedical ontologies and patient data. In our method, we output answers to a list of clinically relevant questions on a per patient basis, to provide contexts around entities of interest in the risk prediction of comorbodities of chronic disease(s). Given the interdisciplinary nature of our work, in that we utilize computational methods to aid a clinical use case like comorbodity risk prediction, we have also used user-driven strategies like consulting with a medical expert during the development of our methods and conducting an expert panel session to improve upon its domain relevance. Additionally, since the comorbidity risk prediction problem is broad, we have chosen to focus our approach on predicting chronic kidney disease, one of the major comorbidities of type-2 diabetes. 

In this section we analyze both the quantitative and qualitative results presented in Sec.~\ref{sec:results}. We analyze both the feasibility of supporting contextual explanations from authoritative sources such as CPGs, and the usefulness of providing contextual explanations from an analysis of the themes derived from our expert panel sessions.

\subsection{Feasibility of extracting and generating contextual explanations from
authoritative sources}
For our feasibility analysis, we further analyze the results for type 3 questions with respect to the disease groups to gain a deeper understanding of the QA performance. 
Specifically, we want to understand whether SOTA LLM backed QA methods, potentially augmented with knowledge, are ready for real-world use for our states use-case as well as identify patterns that might apply to other CPGs in different disease areas. We pose a number of questions around this idea as follows:

\noindent \ul{Is CPG guideline suitable for {\dm} contexts}: \textit{How well are the disease subgroups among {\dm} patients covered in the guidelines?}
\begin{figure*}[!htbp]
    \centering
    \includegraphics[width=\linewidth]{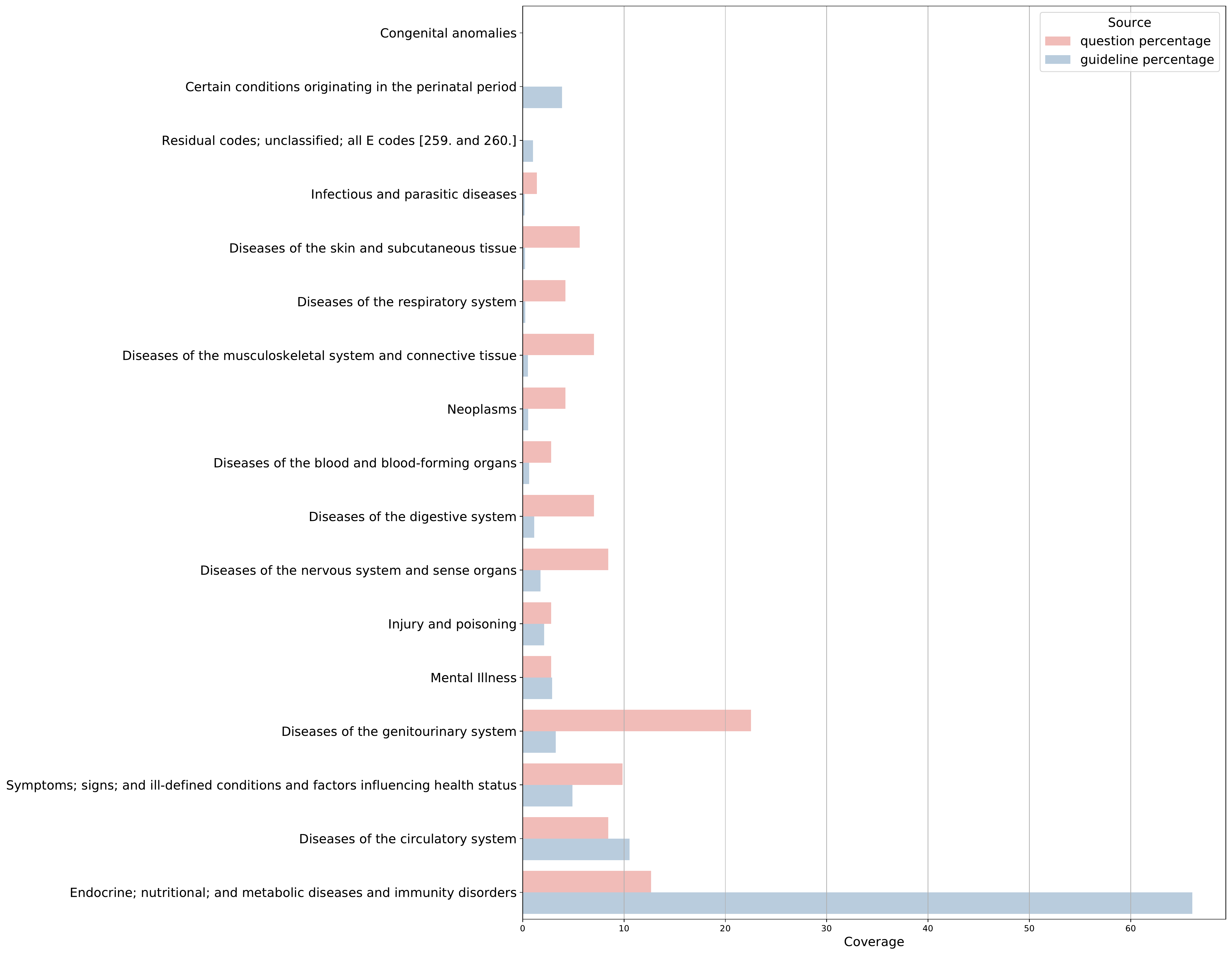}
    \vspace{-2em}
        \caption{Comparing disease group occurrences in the ADA CPG 2021 versus those in the feature importance questions from our chosen prototypical patients.}
        \label{fig:guidelines_patient_diseases}
\end{figure*}
Here, we attempt to understand the applicability of ADA 2021 CPG in our use-case. From Fig.~\ref{fig:guidelines_patient_diseases}, we can interpret that the guidelines cover a smaller number of disease groups~\footnote{Disease groups are derived by rolling up disease codes both in the patient data and guidelines to their higher-level CCS LVL 1 groups.} than the patient data. Since CPGs are authoritative literature in their disease fields, their coverage is mainly limited to the primary disease area. Thus ADA 2021 CPG focuses on Diabetes, an Endocrine, Nutritional and Metabolic Disorder, and its comorbid conditions (mainly spanning diseases of the circulatory and genitourinary systems). 
Unsurprisingly, these patterns are seen in the Fig~\ref{fig:guidelines_patient_diseases} as well where the Endocrine, Nutritional and Metabolic Disorder have the largest coverage in the guideline data, a $66\%$.
In contrast, patients might have other conditions that do not arise from the {\dm} diagnosis alone, and hence we can deduce that we see more diversity in disease groups in the patient data. 
%Next we attempt to answer if any of the LLMs favor some disease groups over the other.

\begin{figure*}[!htbp]
    \centering
    \includegraphics[width=\linewidth]{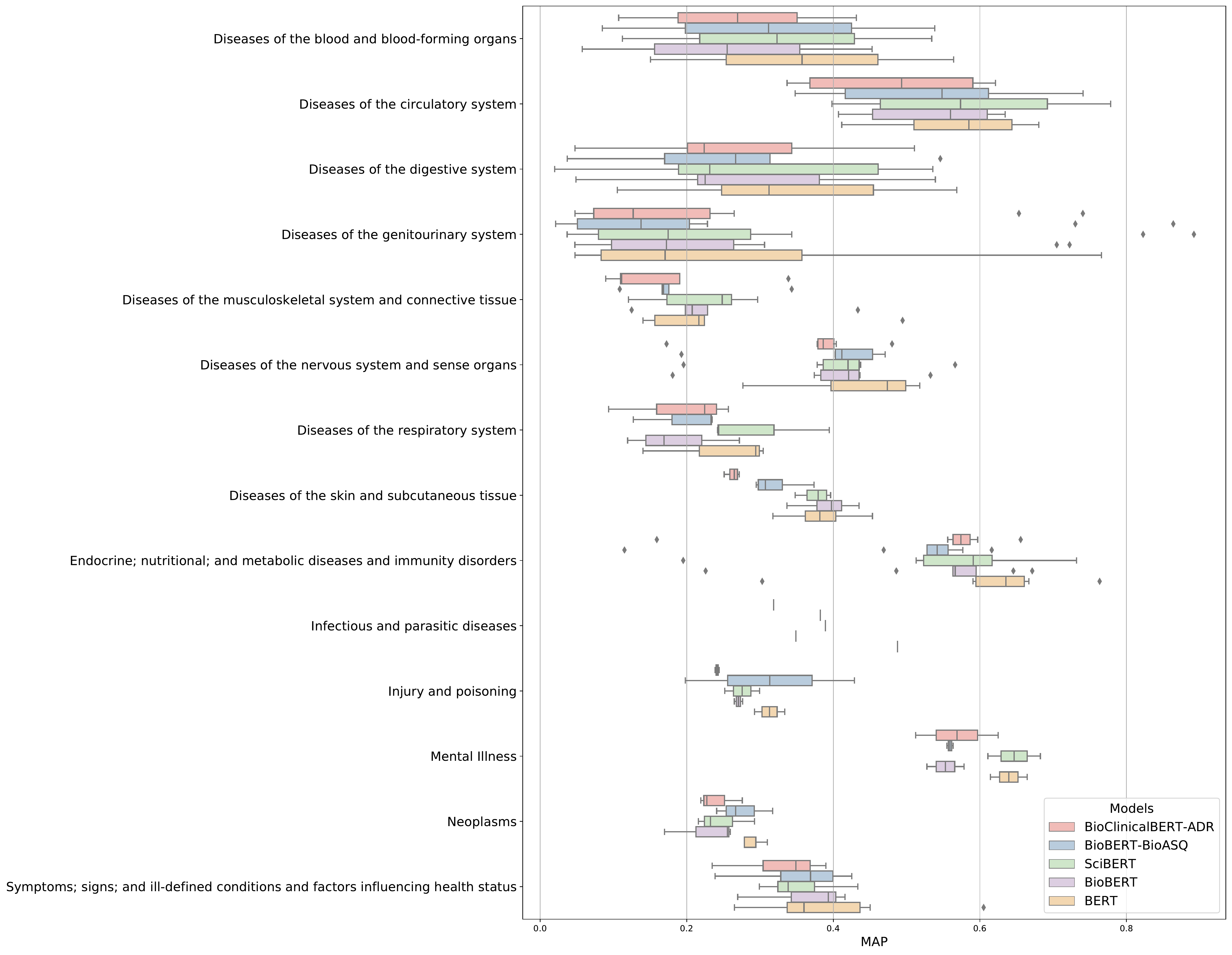}
    \vspace{-2em}
        \caption{Overall view of the performance of all language models in our experiments over the $14$ disease groups covered in feature importance questions.}
        \label{fig:model_subgroup}
\end{figure*}

\begin{figure*}[!htbp]
    \centering
    \includegraphics[width=\linewidth]{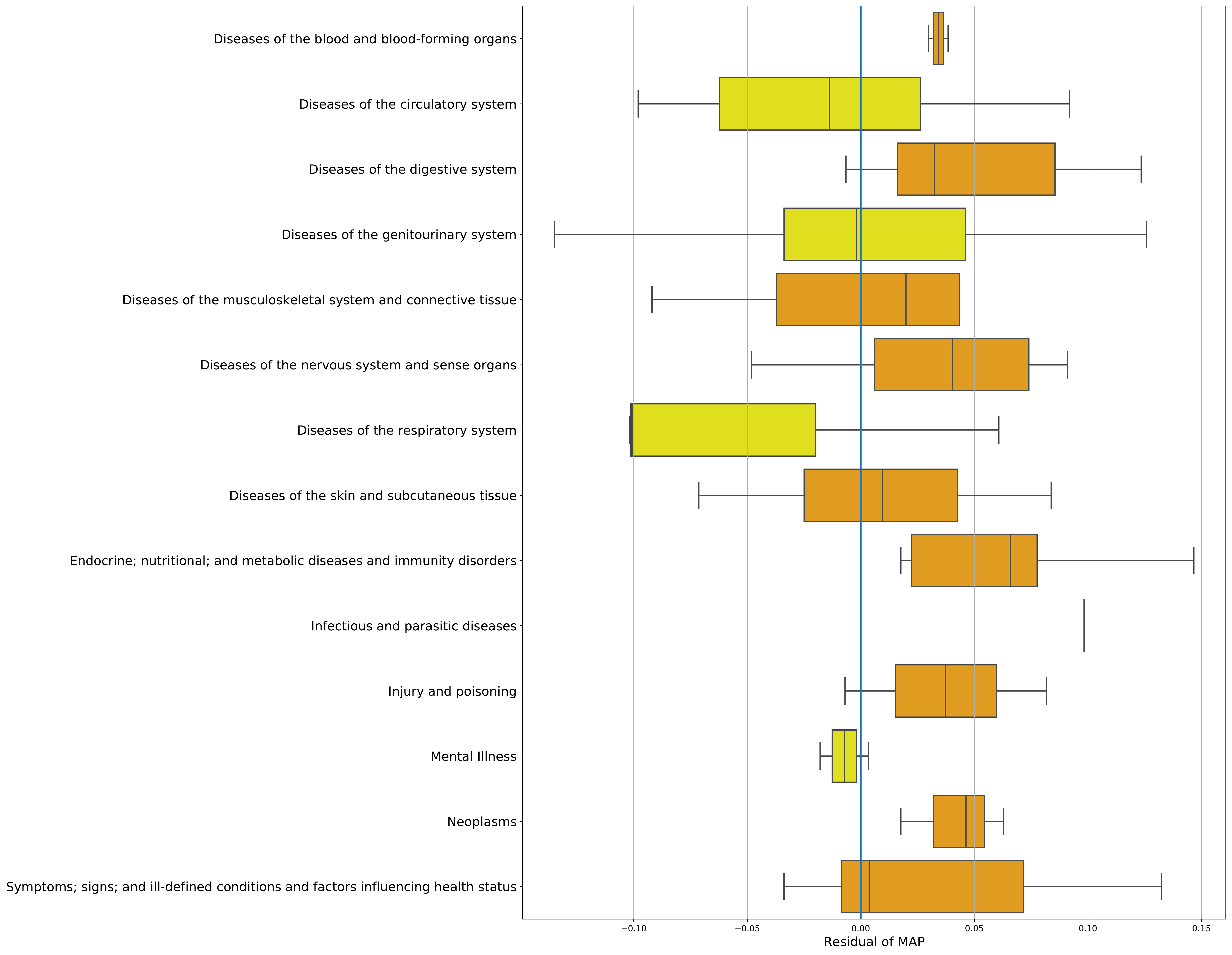}
    \vspace{-2em}
        \caption{Comparative performances of top-performing native language models, BERT vs. SciBERT. Plotting the residuals under equal performance hypothesis for the $140$ feature importance questions that span $14$ disease groups. \textcolor{orange}{Orange} box indicates BERT performs better on average while \textcolor{yellow}{Yellow} indicates SciBERT is the better choice for the disease group.}
        \label{fig:bert_scibert_boxplot}
\end{figure*}

\noindent \ul{Can a single SOTA LLM method be used to extract the contexts?}
We attempt to understand if the LLMs are inherently better at certain disease groups over others. Fig~\ref{fig:model_subgroup} shows the distribution of base LLM models over the disease groups. We can see that there are a few disease groups which have a higher MAP performance than others (towards the right end of the plot), some have their the box centers in the middle of the plot and others who are not doing as well since they are in the first quadrant of the plot. While the results are not strikingly decisive and statistically significant everywhere,  in concordance with our overall results, we note that SciBERT and BERT models have  better performance over most of the disease groups.
%since their boxes are generally ahead of the other model boxes.
Thus to further discern between these top $2$ performing models, we conducted a point-wise analysis of relative performance difference between BERT and SciBERT (distribution of residual values between the MAP performances of BERT and Scibert under equal performance hypothesis). Fig. \ref{fig:bert_scibert_boxplot} shows the outcomes of the analysis where orange box indicates that BERT performs better on average while yellow indicates the same for SciBERT. 
We see that BERT is better for most disease groups, especially for `Disease of the blood and bone forming organs', `Diseases of the digestive system', `Diseases of the nervous system and sense organs', `Endocrine, nutritional, and metabolic diseases and immunity disorders', `Injury and poisoning', and `Neoplasms' ($0$ not contained in the inter-quartile range).   SciBERT is only doing better on `Diseases of the respiratory system' (and marginally better for `Mental Illness'). %two groups (to the left of the zero line). 
These results, in addition to the quantitative results, indicate that LLM models are better at addressing some disease groups than others. While vanilla BERT is a defensible choice, the results point to the need for domain adaptation for LLM for this problem. However, considering the limited availability of data, novel ML methods such as one-short learning and as weak supervised techniques may be 
required to improve the performances of these LLMs reliably across multiple disease groups. 
\vspace{1em}

\begin{figure*}[!htbp]
    \centering
    \includegraphics[width=\linewidth]{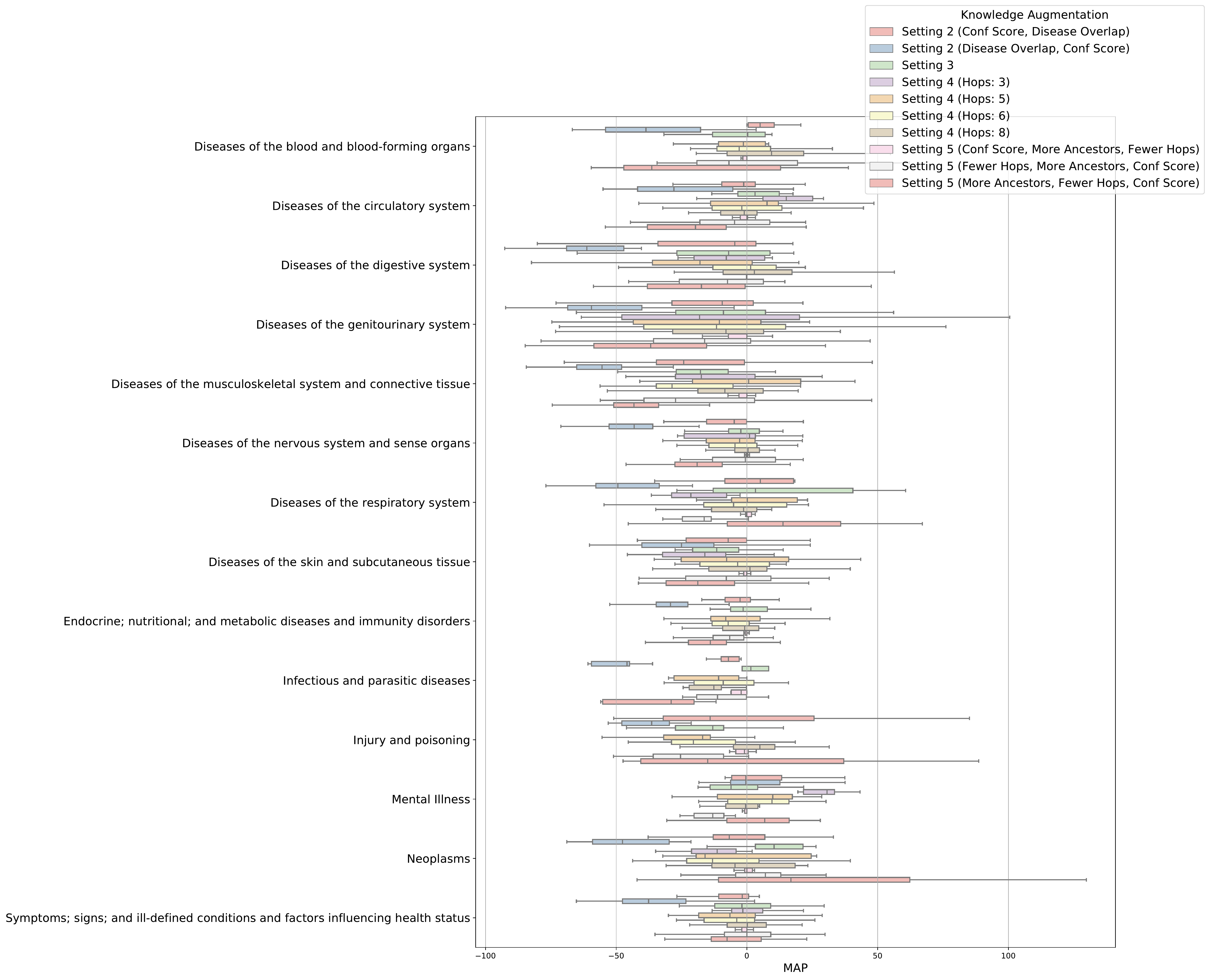}
    \vspace{-2em}
        \caption{Lift in performance over native language models by different knowledge augmentation strategies over the $14$ disease groups covered in feature importance questions.}
        \label{fig:knowledge_subgroup}
\end{figure*}
\noindent \ul{Does knowledge augmentation reliably improve QA methods?}
We proposed $4$ possible strategies for Knowledge-augmentation (See Appending~\ref{sec:qamethod:appendix}).  Among these settings, the best knowledge augmentation strategies reflected in Table.~\ref{tab:kmdiseaseresults} originated from a composite of strategies. As seen from Tab.~\ref{tab:nativemodelresults}, ~\ref{tab:kmdiseaseresults} and ~\ref{tab:drugresults}, the guideline QA's \textit{best MAP score of 0.82} is obtained in a BERT + knowledge augmented setting on drug questions and among disease features, the guideline QA's \textit{best MAP score is 0.438}. The \textit{recall with its highest value of 0.405} is obtained in a post-filtering knowledge augmentation setting 5 of SciBERT (Tab. \ref{tab:kmdiseaseresults}), where we sort answers by disease overlap between question and answer and we also see the best \textit{BLEU score of 0.19} in this setting.
These points to the fact that there is value in either filtering the answers to be passed to a LLM  or sorting the answers from it, using aids from known domain knowledge sources that data sources like guidelines are expected to adhere to.
We further analyzes these at disease sub-group level in Fig.~\ref{fig:knowledge_subgroup} where we plot the lift in performance over corresponding base LLM using any particular strategy. While any one strategy is not found to be dominating the others, for most disease groups we can find one or more strategy that improves the performance (median lift greater than $0$). These results support our previous insight that there is a value in augmenting domain knowledge. However, finding a single universal strategy is difficult and may need further research.

\vspace{1em}
\noindent \ul{Overall, how feasible it is to extracting contexts from guidelines? Which strategies are beneficial? Are the methods scalable?} 
% PC Last question: what are the key take-aways? Can this be scaled?
We have addressed $175$ clinically relevant questions that provide context around $20$ prototypical patients, their predicted risk, and the factors influencing their risk. We have implemented logical adaptations given what we know about the guideline data to improve the LLM model's capabilities and performance. These adaptations include knowledge augmentation from well-used medical ontologies like Metamap and Snomed to improve semantic overlap and rule augmentation to address numerical range questions. Our baseline LLM, BERT itself, has a variable performance that does well on some questions and not on others. 
% The performance of BERT on $71$ disease feature questions and in comparison to our annotated dataset of the ADA 2021 CPG is a F1 score of $0.21$, precision of $0.39$, recall of $0.22$, and BLEU score of $0.12$.
Similarly, the best performances on the LLM + knowledge augmentation approaches varies across pre-filtering settings 3 and 4, that filter by Metamap disease codes and Snomed disease hops and post filtering-setting 5 that sorts by Snomed disease hops. 
 From our result evaluations, we see that the order of introducing the knowledge augmentation outputs impacts the accuracy scores, namely the MAP and recall. Mainly, pre-filtering the answer set before passing to a LLM can help it output more precise answers. In the best case, pre-filtering settings provide a gain of $4\%$ over the baseline LLMs both for disease and drug questions (BERT-KA from Tab. \ref{tab:nativemodelresults} and BERT-KA from Tab. \ref{tab:drugresults}). Similarly, post sorting the answers from a LLM can improve the recall, and in the best case (SciBERT-KA from Tab. \ref{tab:kmdiseaseresults}), we see a gain of $5\%$ from the baseline LLM. 
 
Our result numbers also indicate that unsupervised adaptations can only reach a certain accuracy and point to the need for domain adaptations to medical guidelines. Additionally, since we were dealing with a setting with little or no annotations on the ADA 2021 CPG, we had to create our own annotations. Currently, we are dealing with a relatively small annotated corpora ($85$ questions and $654$ candidate sentences), and we consulted with a medical expert on our team to review these annotations. Even for this small corpora, we find that it is time-consuming for a clinical expert (s) to review the annotations or create them. We are exploring techniques like weak supervision to scale and improve the coverage of the annotations. In summary, our guideline QA results depict incremental gains in adding knowledge and rule augmentations to enhance LLMs' performance and capabilities in domain applications and point to the need for supervised and semi-supervised approaches to improve these gains.

We are, to the best of our knowledge, the first to report any QA performance numbers on the ADA CPG 2021 dataset. 
% \todotip[SC]{Add annotations here.}
Additionally, we are the first few who have tried a LLM approach for more scalable upstream tasks on medical guidelines like question answering than the current more time-consuming and dataset-dependent task of converting guidelines to rules and applying logical reasoning techniques over these rules~\cite{gatta2019clinical, riano2019ten}.
Our approach to guideline extraction and question answering (Fig. \ref{fig:guidelinearchitecture}) is a step towards providing a more flexible way (~\cite{hematialam2021identifying, hussain2021text}) to swap in guideline text from different diseases as needed. Our enhanced LLM approach (Fig.~\ref{fig:guidelinearchitecture}) can be applied to any medical text corpus like medical guidelines extracted into a machine-comprehensible format and can address different question types (as seen in Tab.~\ref{tab:questiontypes}) relevant in risk prediction settings.

\subsection{Understanding the added benefit of the derived contexts}

\noindent \ul{What were the takeaways and feedback from clinicians about the supported contextual explanations?} The four major themes - \textit{Clinical Value of Explanations and Contextualizations, Highlighting Actionability, Connections to Patient Data, and Connections to External Knowledge Sources} - that we found during the expert panel interviews to evaluate our contextualization approach, mainly point to the overall value of supporting different types of contexts, both from literature and patient data, and the need to better present connections between these contexts. Many of the contexts the clinicians on our expert panel were looking for were around the post-hoc explanations of the factors contributing to the risk. This finding corroborates a recent study that reports that post-hoc explanations themselves are insufficient to provide reasoning that clinicians can interpret and act upon~\cite{GHASSEMI2021e745}, and also add to the well-accepted belief that risk scores are insufficient. 

\vspace{1em}
\begin{figure*}[!htbp]
\centering
\includegraphics[trim=0 0 0 30,width=\linewidth]{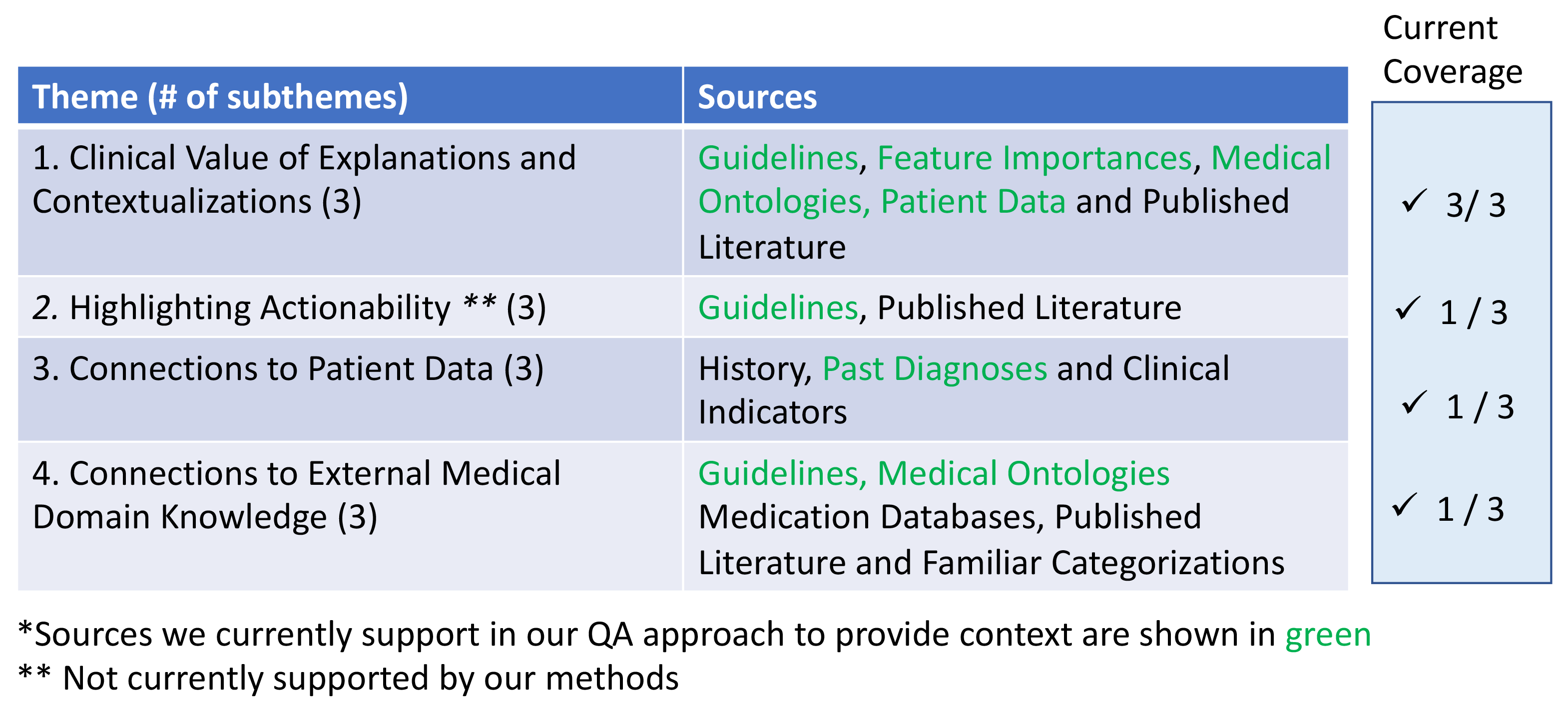}
\caption{Summarizing the coverage of our current data sources to support the themes we found from our expert panel discussions.}

\label{fig:themecoverage}
\end{figure*}
\noindent \ul{Do the supported data sources address the clinicians needs?} 
Through further analysis, we find that the contexts the clinicians were looking for and discussing can be addressed either by connections to patient history and data - patients' diagnoses, medications, and lab values - or through published literature. Specifically, we find that the different questions and question types (Tab. \ref{tab:questiontypes})
that we support from the {\dm} guidelines can address \textbf{$6$ of the $12$ sub-themes} (Fig. \ref{fig:themecoverage}), i.e., providing contextual information around patient's {\dm} state, their {\ckd} risk and the individual features (Theme 1), highlighting the impact of {\ckd} risk on treatment decisions for {\dm} (Theme 2), providing links to published articles (Theme 4), and showcasing connections to patient clinical indicators (Theme 3) where mentioned. Some other themes can be easily addressed by enabling connections from the {\ckd} risk scores and the features contributing to them, to patient timelines for diagnoses and lab values. Other themes - support for familiar categorizations (Theme 4) and the need for information on related diagnoses (Theme 3) - benefit from connections to medical ontologies that support either drill-downs to more specific diagnoses or abstracting up to higher-level pathways. We currently only support abstractions to higher level physiological pathways on the prototype dashboard (e.g., all disease of the circulatory system can be filtered from the patient's feature importances) and are investigating how to support drill-downs based off of these pathways more broadly.

% We are continuing to build methods to improve the overall accuracy and coverage of our current guideline QA approach.  We next highlight the added value of building processing capabilities around authoritative literature sources like medical guidelines and doing so from clinically relevant perspectives.
%SC Can add if reviewers ask
% and the post-hoc explanation method, SHAP~\cite{lundberg2017unified}, had a faithfulness score of $Z\%$.
\begin{figure*}[!htbp]
\centering
\includegraphics[trim=0 0 0 30,width=\linewidth]{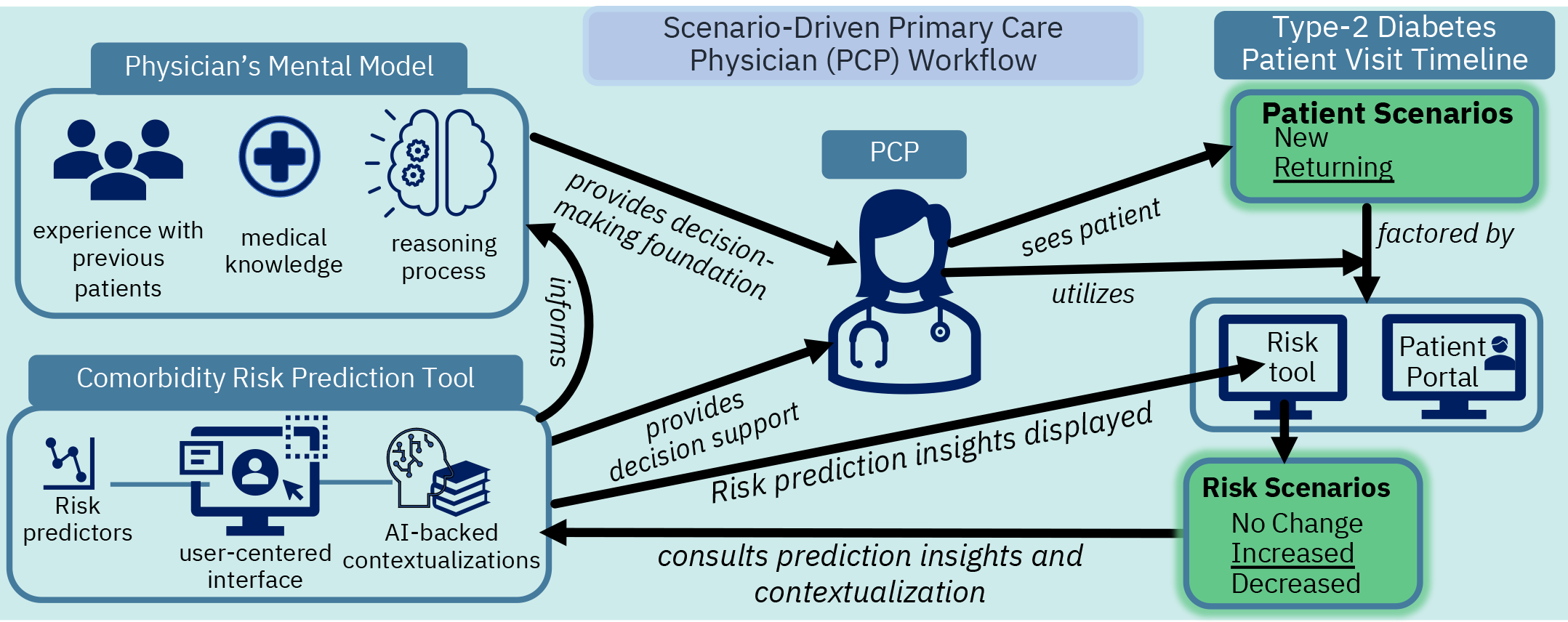}
\caption{Illustration of findings for a Primary Care Physician (PCP) - one target persona among clinicians - whose workflow is dependent on the patient context and the clinician-patient history. We show an example scenario where the predicted risk for a returning patient has increased since the last visit. This context and the PCP's mental model drive the PCP's following actions, such as differential treatment decisions made by probing reasons for the increased risk.}

\label{fig:PCP-workflow-diagram-condensed}
\end{figure*}
\noindent \ul{What can be the impact of our contextual explanations beyond the comorbidity risk prediction setting?} 
To identify specific scenarios in which our contextualizations might be most impactful, we discussed with a clinical expert typical situations of {\dm} patient care by clinician type and patient characteristics  to understand where risk predictions %and better forecasting 
could help clinicians improve patient care planning. 
While current literature~\cite{wang2019designing} on designing healthcare models points to a user-centered approach, from this understanding of clinician workflows, our discussion showed the importance of grounding such user centered work in specific clinical scenarios (see Fig~\ref{fig:PCP-workflow-diagram-condensed}). 
For example, it became clear that dashboards containing contextual explanations around risk prediction could be used by clinicians in different ways. For example, a clinician seeing a patient for the first time and/or for the first diagnosis would be interested in creating their mental model of the patient's diagnosis and understanding the causes for the risk, whereas a clinician seeing a returning patient with a previously established diagnosis (where clinicians might be more interested in understanding any changes to the  risk prediction over time and the effectiveness of various interventions). 
Hence, based on these understandings, we formalized our use case to provide contextual explanations to a PCP around the predicted risk of {\ckd} among new {\dm} patients at their first diagnosis.

Additionally, while we focus our approach in the risk prediction of {\ckd} among {\dm} patients, the contextualization approach can be applied to other comorbid risk prediction settings given access to authoritative guidelines in the disease area, and likewise, the themes that we analyze from our expert panel interviews are also general enough to be considered applicable in other disease settings. These themes indicate a larger need for AI systems to support insights from multiple data and knowledge sources and present them as actionable and contextual explanations~\cite{chari2020explanation} (also pointed out in the self-explanation scorecard from~\cite{mueller2021principles}). In summary, our approach is a step towards extracting clinically relevant context from different data and knowledge sources, including guidelines, patient data, and medical ontologies, and using these contexts to augment and explain answers to a list of clinically relevant questions that can help clinicians reason and interpret the risk predictions for patients.

% In the following sections, we describe the constructed risk models and structured, scenario-based interviews with an expert panel to uncover the contextual dimensions that will enable them to use the AI models in practice.

% We engaged with experts at every stage of the development lifecycle, from formalizing the use case, designing features in the pipeline that would support our use case from a clinical standpoint, and finally, evaluating our prototype risk prediction dashboard and its supported features via expert panel interviews. This ensured that our approach was validated at each stage in hopes that the features we support to provide context in a comorbidity risk prediction setting provide the ``most effective amount of information to provide to users,'' are provided how and when clinicians need it, and are conveying the model’s confidence in its insights correctly~\cite{park2020evaluating}. 

\vspace{1em}
\noindent \ul{What are some future directions that emerge from the clinician discussions?} 
Some subthemes under \textit{Theme 2: Highlighting Actionability} provide future directions, highlighting actionable factors and suggesting specific actions to reduce {\ckd} risk are not currently addressed by our contextualization approach. These sub-themes require more investigation and development of methods to identify actionable, most relevant factors to {\ckd} risk. Another point which we observed is that some of the factors that the model picked up on are not covered by the {\dm} guidelines, and could either be factors only relevant to {\ckd}, or are those that are not considered to be well-known enough to be covered in position statements like CPGs. We are also investigating how to combine insights from multiple guidelines (also mentioned in~\cite{sittig2008grand}) and if that would be useful. In summary, while some of these themes provide validation for the modules we currently support in our multi-method approach to provide context, others offer directions for us to build towards, such as enabling connections to external medication databases, supporting temporality in post-hoc explanations of risk,
%SC cite the paper from KDD speaker 
%SC mention our plan to use some rule of them to identify actionable and modifiable factors - cite papers with meta understanding of factors 
and efforts to better present answers in terms of relevance and actionability. We are also considering interviewing more user groups within the clinical domain to strengthen an understanding of where such a risk prediction tool would be most impactful. 
Future steps would also include a practical study in a clinical setting to further assess the utility of our method.

%Prompt engineering, since our questions are not patient specific.

\section{Related Work}
\label{sec:relatedwork}
% \todotip[SC]{Dropping in workshop paper version. Move it after discussion. Keep it close to contributions. Diff. viewpoints - risk predictions, explanations for risk predictions, user-driven XAI and survey paper on contextual explanations.}

Our methods %aims to build 
build on both expert feedback and past efforts to leverage clinical domain knowledge for generating explanations within AI assistants. 
Some %of the notable and most relevant ones are 
notable and relevant past works include: MYCIN~\cite{shortliffe1974mycin}, where domain literature was encoded as rules and trace-based explanations, which addressed `Why,' `What,' and `How,' were provided for the treatment of infectious diseases; the DESIREE project~\cite{seroussi2018implementing}, where case, experience, and guideline-based knowledge was used to generate insights relevant to patient cases; and a mortality risk prediction effort~\cite{raghu2021learning} of cardiovascular patients, where a probabilistic model was utilized to combine insights from patient features, and
% supporting evidence from 
domain knowledge, to ascertain patient conformance to the literature. However, these approaches are either not flexible nor scalable for the ingestion of new domain knowledge~\cite{shortliffe1974mycin, seroussi2018implementing}, or are narrowly focused in their approach to explanations along limited dimensions~\cite{raghu2021learning}. 
%Through our approach, w
We attempt to allow clinicians to probe the supporting evidence systematically and thoroughly while asking holistic 
questions about the supporting evidence(s) 
% (both from the literature and patient cohort) 
to understand their patients better.

%\todotip[SC]{Add more content around LLLMs for biological data.}

On the guideline QA front, there have been several efforts on representation formats
%Cite GLIF and GATE - the pubmed review 
for guidelines and more recent work on applying machine learning and language model approaches on guidelines for upstream tasks other than QA~\cite{hussain2021text,hematialam2021identifying, schlegel2019clinical}. Guideline representation efforts attempt to model guidelines as rules that can then be checked against patient data for conformance. While rule engineering is more accurate than applying machine learning models, it is not scalable without human effort. In a more scalable effort, Schlegel et al.~\cite{schlegel2019clinical}, have shown how a standard NLP pipeline of tools like named entity recognizers, syntactic, semantic, and dependency parsers can be applied to convert guideline text into annotated text snippets. However, their system, ClinicalTractor, is not available for reuse yet, and hence we could not use it for the semantic annotation portion of our QA pipeline. Another similar effort is from Hussain et al.~\cite{hussain2021text}, where they use heuristic patterns to identify different composition patterns in guideline sentences. These guideline natural language understanding efforts while useful, still require significant effort to be used upstream by QA approaches and could instead be used to augment QA approaches such as ours with additional information that can help improve semantic and syntactic coherence of answers.

% \todotip[Elif | PC | MG]{Defend how NLU can be compared to QA}

%SC mention that these ontologies have been used in QA systems 
%SC a couple of QA papers doing guideline NLP then it is fine - supported by community 
On the other hand, with the rise of LLM~\cite{devlin2018bert}, several papers have been published on adapting language models to the biomedical and clinical domains by the pre-training of these models on large biomedical corpora~\cite{lee2020biobert, gu2021domain}. There are currently very few efforts on the applications of these domain adapted language models to medical guidelines~\cite{hematialam2021identifying}. Hussein and Woldek~\cite{hematialam2021identifying} have applied language models to identify condition-action statements from three medical guidelines, and they find that the combination of syntactic and semantic features from Metamap can boost the performance of language models like BioBERT. However, it is not immediately clear how the extraction of condition-action statements can be used in a QA setting where the range of question types like those we support goes beyond condition-action pairs. For example, questions asking about diagnoses features don't always have a condition to be searched against.  Contrarily, Sarrouti et al.~\cite{sarrouti2020sembionlqa}, have designed a semantic biomedical question answering system that achieves the state-of-the-art results on the BioASQ challenge by using UMLS similarity scores and a novel passage retrieval algorithm to find candidate answers from Pubmed documents. In their future work, they list the needs for large training samples as a limiting factor to use a deep learning algorithm. While we agree, we have shown how knowledge augmentation algorithms can improve the performance of deep learning language models in new settings such as unseen guidelines.

Several studies have also tried to utilize patient data to query the literature for applicable evidence or treatment suggestions. However, very few of these studies combine multiple modalities and sources of data and knowledge for querying. 
Agosti et al.~\cite{agosti2019analysis}, conducted an analysis of query reformulation techniques for precision medicine.
Natarajan et al.~\cite{natarajan2010analysis}, conducted an analysis of clinical queries in an electronic health record search utility and found that queries on diseases and lab results were most searched.
 Patel et al.~\cite{patel2007matching}, matched patient records to clinical trials using ontologies and used a purely logical A-box and T-box approach to query literature.

%SC Need a concluding sentence here 

Finally, several studies have hinted that model explanations alone are not sufficient and indicate that context can be an important dimension to make these explanations more useful. We summarize a few studies which either contextualize model explanations with links to knowledge or can provide context around risk prediction scores to make them more useful. 
 Rieger et al.~\cite{rieger2020interpretations}, found that interpretations are useful by penalizing explanations to align neural networks with prior knowledge.
 Zhang et al.~\cite{zhang2021context}, presented context-aware and time-aware attention-based model for Disease Risk Prediction with Interpretability. They used disease code hierarchies as context in RNN network's attention layer. Weber et al.~\cite{weber2021knowledge}, attempted a Knowledge-based XAI through CBR and found that there is more to explanations than models can tell. Yao et al.~\cite{yao2021refining}, refined Language Models with Compositional Explanations by align LLM and post-hoc output with human knowledge and Tonekaboni et al.~\cite{tonekaboni2019clinicians}, analyzed what clinicians want and found that Contextualizing Explainable Machine Learning for Clinical End Use.
%Pointers for discussion content
% Flow should be user study identified themes - some are covered in guidelines - what are limits of extraction (guideline QA performance) -- mention drug database integration in future -- find sequence of answers 

\section{Conclusion} \label{sec:conclusion}

Contextual explanations have been posited to be useful for clinicians for real-world usage of AI models. In this paper, we have developed an end-to-end AI systems and studied the feasibility and usability of extracted contextual explanations from medical guidelines using state-of-the art QA methods. 
We have focused our study in a risk prediction use-case for {\ckd} among {\dm} patients and have conducted both quantitative and qualitative analysis.
%We have designed a multi-method approach to provide \textit{context or additional information to connect} entities of interest in the risk prediction setting of a type-2 diabetes comorbidity, chronic kidney disease, 
%to relevant knowledge from multiple clinical sources. 
Upon conversations with clinicians, we have selected three entities of interest in the risk prediction setting to provide contextual explanations along - the patient, their predicted risk, and the model explanations of their risk. 
%We have also shown how a comorbidity risk prediction tool such as ours would fit in the larger clinical workflow. 
%As a second contribution, we have developed multiple methods to extract and include contextual information from patient data, medical ontologies, and clinical practice guidelines to answer a list of clinically relevant questions. Finally, we have evaluated the usefulness of a prototype risk prediction dashboard that embeds our generated contexts with an expert panel of clinicians. We have presented and discussed the $4$ higher-level themes and $12$ sub-themes that arose from an analysis of these interviews and depict how we are addressing these themes. 
Crucially, we have identified several themes covering the explainability needs of clinicians. The supported contextual explanations support some of these themes and thus improves clinician's confidence is using AI supported systems. We also found state-of-the art large language models to be effective in extracting such contexts, especially for certain disease sub-groups. While our results support the hypothesis that presenting contextual explanations to clinicians is both feasible and usable, the performance and requirement gaps points to the need for further research in this field.
%In addition to the thematic analysis of the expert panel results, we have provided quantitative performance metrics for the risk prediction models and our question-answering (QA) modules. 
For example, while we have considered three domain sources for the contexts in this paper, the themes from the expert panel interviews also indicate that there may be value to connecting to other sources, including extracting additional guideline details from tables and flowcharts, and potentially
%check if this is mentioned in intro / data sources
involving multiple layers  of the evidence pyramid to include such sources as systematic reviews, randomized clinical trials, cohort studies, and expert opinions. 
% \todotip[SC]{Add a sentence on guideline QA performance.} 
%SC Done
%We have created a small annotation dataset for systematic evaluation of our novel, unsupervised guideline QA and our results across the different knowledge and knowledge augmented settings of the QA approach indicate that there are research opportunities to improve accuracies in authoritative clinical literature.  
% small dataset
% annotations that we created to evaluate the performance systematically
% unsupervised methods used
Similarly, novel machine learning techniques such as weak-supervision or one-shot learning may be need to improve the quality of extracted contextual explanations. A combination of both may also enable other approaches such as `prompt engineering' whereby patient data can be used to seed the QA model questions and get richer response.
Our future research will be directed at overcoming the aforementioned opportunities. Overall, by closely working with clinical experts and adopting inter-disciplinary approaches, from the use case crystallization and methods development, to the evaluation stages, we have shown the promise in supporting clinically relevant contexts to help clinicians better interpret risk prediction scores and their model explanations.

% \begin{itemize}
%     \item Context helps and is needed to explain explanations --> our user study supports more explanations are necessary; broadly liked the idea.
%     \item The guidelines contain the most updated systematic evidence in the field, and surfacing evidence from such resources are helpful --> 
%     \item We have a working implementation and are able to support multiple forms of contextualizations %SC move to conclusion --> 
%     \item Conducted a study with experts and they are interested about certain themes 
%     \item What can support these contexts and how? Connect it to the scientific pyramid; guideline give 30\% of theme A
%     \item If guidelines can support A, are we able to extract?  
% \end{itemize}
% risk prediction numbers + factors are not easily digestible by clinicians + what else is needed we have done a thematic analysis + novelty in implementation (symbolic-based QA method to extract answers from guidelines) + overall found guidelines satisfy X, Y needs and our method have performace, and we can expand to other modalities or go down scientific pyramid has value.

\section*{Acknowledgments}
This work is supported by IBM Research AI through the AI Horizons Network. 
%SC worst case we remove this next sentence
We also thank the clinicians on our expert panel discussions.
%Dr. Mehool Patel, Dr. Phillip M. Francis, Dr. Rubina Rizvi, and Dr. Jeffrey T. Lenert for their expert feedback. 
We thank Rebecca Cowan from RPI and Ching-Hua Chen from IBM Research for their helpful feedback on the work.

 \bibliographystyle{elsarticle-num} 
 \bibliography{main}

%% The Appendices part is started with the command \appendix;
%% appendix sections are then done as normal sections
\appendix 
\section{QA Architecture} \label{appendix:qa_architecture}
We discuss in this section the extraction methods for information retrieval from the different data sources that we support as context, the sub-modules which help in question and explanation generation, and finally, the standard evaluation modules which can output scores for some of the question types. 
\subsection{Information Retrieval} \label{sec:informationretrieval}
%SC for performance report the IR error in the BeautifulSoup code in the appendix 
%SC add manual augmentation part here due to variances in that the files were structured 
Here we describe the information extraction portion of our QA pipeline, as seen in part A). of Fig. \ref{fig:guidelinearchitecture}. 
We support the extraction of context from three domain sources in our QA approach, including patient data, medical ontologies like Clinical Classification Software (CCS) codes~\footnote{\url{https://www.hcup-us.ahrq.gov/toolssoftware/ccs10/ccs10.jsp}} and medical guidelines from ADA Standards of Care 2021 (as introduced in Sec. \ref{sec:background}). We query patient data from Limited Claims Explorys Dataset (LCED) claim records (see Sec. \ref{sec:use-case}) on-demand, either when we need to create questions based on patient parameters or when we need to include these patient values in answers to questions about the patient. As for extracting content from CCS codes, we download a static version of the `CCS for ICD-10-PCS Tool' file and use four fields from this downloaded file - `Field 2: CCS Category,
Field 3: Code Description, Field 7: Multi-level 2 Category, and 
Field 8: Multi-level 2 Category Description'. Field 8 in particular, connects the lower level disease codes we see in the patient data to their higher level 2 groupings (e.g., Essential Hypertension's Multi-level 2 Category Description is Disease of the Circulatory System). 

 We extract content from the HTML or web version of the `Standards of Medical Care in Diabetes'~\cite{care2021standards} guidelines, published by the American Diabetes Association (ADA)~\footnote{\url{https://care.diabetesjournals.org/content/44/Supplement_1}} using a Python library, BeautifulSoup~\cite{richardson2007beautiful}. The ADA CPGs, are updated annually and are an example of a well-maintained CPG, whose format has been fairly consistent over the past decade. The ADA CPG is released both in web-friendly formats like HTML and in PDF formats. The content in the ADA CPG is split across chapters, where each chapter focuses on a different aspect of diabetes management or treatment. Furthermore, each chapter contains different recommendation groups, each containing recommendations, discussions, and references for a sub-area within the chapter (Fig. \ref{fig:guideline-structure})~\cite{chari2019making}. The recommendations themselves are supported by different grades of evidence and are graded accordingly. For example, a recommendation supported by a systematic review or meta-analysis is assigned a grade A, whereas only an expert opinion is graded as E. Also, while the references for the recommendations are not made available as direct associations, they can be found within the discussion supporting the recommendations. Mainly, within each chapter in the ADA guidelines, we extract the content of different sections, including the recommendations, supporting discussions, and references within these sections. We then write the output of this extraction, mirroring the structure of the original guidelines (see Fig. \ref{fig:guideline-structure}), to a semi-structured JSON format, which is then used within our QA modules.

\iffalse
\begin{figure*}[!htbp]
    \centering
    \includegraphics[width=\linewidth]{images/ebm.jpg}
        \caption{A visual representation of the evidence-based pyramid in the clinical domain, taken from Rosner, 2012~\cite{rosner2012evidence}. As seen at the top of the pyramid are systematic reviews and meta-analyses, and the clinical practice guidelines can thought to be systematic reviews often summarizing latest evidence from the disease area including evidence from highly cited meta-analyses.}
        \label{fig:ebm}
\end{figure*}
\fi
\subsection{QA Architecture}
\label{sec:qamethod:appendix}

Here we describe the processes and modules, seen in part B) of Fig. \ref{fig:guidelinearchitecture}, that are involved in generating questions based on question types for each patient (see Tab. \ref{tab:questiontypes}) and the answers from the different domain sources of context in our risk prediction setting.

\textit{\textbf{Question Generation}}
The \textit{question generation module} almost always creates templated questions using Python's native support for String Templates,~\footnote{\url{https://docs.python.org/3/library/string.html}} and does so based on patient data, more specifically from the patient's diagnoses codes, lab values, and medication list. The patient's diagnoses codes are sometimes abstracted from their higher-level disease groupings supported by the CCS scheme. A subset of these diagnosis codes can be included in the features that the post-hoc explanation module found were contributing to the patient's predicted risk. In an attempt to provide more context around these features, we create instances of the type 3 question , e.g., ``What can be done for this patient's essential hypertension?'' We also support the creation of two standard, non-variant questions for each patient, i.e., whose values don't change from patient data, that can help clinicians easily interpret their predicted risk (question type 1) and their {\dm} state (question type 2).

Moreover, as can be seen from Tab. \ref{tab:questiontypes}, each of the question types that we support on a per patient basis is populated from different data sources. Hence, we have developed different answering methods for each, including simple lookups and knowledge augmented language model capabilities, including combinations of either a LLM + value range comparison or LLM + knowledge augmentation. We provide examples of questions and answers for each question type in \ref{sec:result_settings:appendix}.

\textit{\textbf{Answer Generation}}
% \todotip[SC]{Start with Answer Generation does so and so}
%SC done
In our answer generation module of our QA approach, we support different submodules that can output answers to questions related to the question types. The answer generation module is capable of inputting questions generated by the previous question generation module and interacting with extracted content from our supported data sources. 

\textit{Template-based Answer Generation:} Question types 1 and 2 from Tab. \ref{tab:questiontypes} can be addressed by simple query lookups of our supported data sources. We populate the Python String Template object with the results of the queried components retrieved by using the widely-used Python Pandas library~\cite{mckinney2011pandas}. This process of creating natural language templates that can then be populated with values on a per-patient basis is supported by the Template-based Answer Generation module of our QA pipeline (Fig. \ref{fig:guidelinearchitecture}). The results of these questions can be summarized or built from structured datasets, like patient data, their model outputs, like risk predictions, and features contributing to their predicted risk and population averages. Hence, there is no fuzziness in the results, which is why we don't evaluate the accuracy of this submodule. 

This submodule is also leveraged in combination with other answer generation submodules when there is a pattern in the answers and slots to be filled. We discuss these details shortly after we set up our knowledge augmented language model capabilities and their usage. 

\textit{Numerical Range Comparison:} BERT LLMs cannot currently determine if a question that has a numerical value comparison, e.g., ``What can be done for patients, whose Hemoglobin A1C $>$ 10?'', falls in the range of the answer returned. However, clinicians often look for recommendations that match patients' lab values in clinical settings such as ours. Further, within the ADA 2021 guidelines, there are mentions for suggestions based on different ranges for lab values, e.g., a recommendation from the Pharmacological Chapter of these guidelines has a recommendation with the mention of ``when A1C levels ($>10\%$ $[$ 86 mmol/mol $]$." Hence, for question type 3 from Tab. \ref{tab:questiontypes}, we need to determine if the patient's lab values are in the range of the answer returned by the LLM module. To address this requirement of performing numerical range comparison between the question and answer produced by LLM, we leverage syntactic parsing capabilities (similar to ~\cite{chen2021personalized}) to identify numerical phrases in the question and answer and then determine if each numerical phrase from the question is in range of the same in the answer. 

We use Natural Language Toolkit (NLTK) chunking and parsing functionalities to identify noun phrases, comparatives, and numerical mentions within both the question and answer. We then write regular expression (regex) rules to identify the patterns of the positional tags returned by NLTK that can constitute numerical phrases. For each of the numerical phrases, we convert them into a tuple of ``(noun phrase, [upper bound, lower bound]).'' This tuple representation allows us to go through the phrases between the question and answer iteratively, and for those that match on the noun phrase dimension, identify if the ranges are in agreement. With these steps, we can then populate an answer using the Template-based Answer generation module, which says if the answer outputted by LLM is in/out of the range of the question. Hence, in this manner, we enhance the capabilities of BERT LLMs for numerical range comparisons via rule-based syntactic methods, which is also why we consider this step a rule augmentation of LLMs' capabilities.

\textit{Knowledge Augmentations to LLMs:} Transformer based LLM approaches like BERT work on sequences of words that are often seen together and their surrounding words, but don't leverage the semantics of whether these words are diseases, medications, or biological processes. We found that in the absence of this semantic knowledge, we would often get answers from the LLM that don't correlate on a semantic level with the question. For example, a sentence from the Comorbidities chapter of the ADA 2021 CPG on Dementia was returned as a valid answer to a question asking about an Abdominal Hernia. To eliminate such answers, we explored options for a biomedical semantic mapper and zeroed in on the National Library of Medicine (NLLM)'s Metamap tool~\cite{aronson2010overview}. We choose Metamap because of its extensive coverage of biomedical semantic types and its ability to capture entity mentions within the ADA 2021 CPG. Within our pipeline, we have integrated a Python wrapper for Metamap\footnote{PyMetamap: \url{https://github.com/AnthonyMRios/pymetamap}} that can recognize biological entities within the guideline text and their semantic types (e.g., dsyn: disease or syndrome, bpoc: biological processes, etc. for a complete list of types returned by Metamap see: \footnote{\url{https://lhncbc.nLLM.nih.gov/ii/tools/MetaMap/Docs/SemanticTypes_2018AB.txt}}).

We run Metamap on question types 3 and 4 from Tab. \ref{tab:questiontypes} to only output answers from BERT when there is a valid semantic match between the question and answer. Specifically, using this knowledge augmentation module, for question type 3, we only output answers whose matched term is a noun and is recognized as a disease term by Metamap, and similarly, for question type 5, we only output answers whose matched term is a noun and is recognized as medication by Metamap. We have observed that depending on the mention of a biological entity in the text, a disease term can be recognized as a disease, biological process, or a finding by Metamap. Hence, we allow for flexibility among semantic types, in filtering disease matches for question type 3. For example, we want to allow answers with the mention of the term `hypertensive' for a question on hypertension, although hypertensive is identified as a finding by Metamap.

Additionally, given this ability to filter based on semantic types, we want to allow additional answers with mentions of related diseases. To provide more broad answers, we use the UMLS Concept Unique Identifier (CUI) codes from the Metamap returned outputs to map to Snomed-CT disease codes~\cite{donnelly2006snomed}. From the mapped Snomed-CT disease codes, we can traverse the Snomed-CT disease tree to identify how many hops apart question and answer disease codes are and if the answer codes are an ancestor of those in the question. We operate on the idea that answers about the parent disease code apply to children nodes. For example, a question about ``What can be done for Asthma'' can borrow from an answer on ``What can be done for respiratory diseases?'' Conversely, if disease codes in the question and answer are far apart in the Snomed tree, it would signify that they are semantically less related. In addition to Metamap codes, we append to the candidate guideline sentences the hop distances from each question computed by applying the Dijkstra's algorithm~\cite{dijkstra1959note} on an uploaded Snomed graph in Python package NetworkX~\cite{hagberg2008exploring} and ancestor values derived from using Python library PyMedTermino's~\cite{lamy2015pymedtermino} is\_ancestor function.

We use the outputs of these knowledge augmentation modules to both pre-filter and post-sort the LLMs answers. The LLM and LLM + post sorting settings 1, 2 and 5, were run against $410$ passage chunks of guideline text, of average length $267$ tokens, since BERT has a 512 token limit for an answer passage. The LLMs on the pre-filtering settings 3 and 5 were run on passage chunks of variable length, depending on the number of filtered sentences to be passed to the LLM model. In the pre-filtering settings, we varied the values of the features that we were filtering by to understand which feature values improve accuracy. In essence, the pre-filtering settings can be thought of as algorithmic knobs to control the set of answers that the LLM has to process. In contrast, in the post-filtering settings we sorted the LLM's answers by feature values, and here we could control the ordering of answers to be outputted. In the pre-filtering setting 2, we filter the guideline sentences by length of disease overlap with the question. In pre-filtering setting 4, we have more possibilities in the feature column because the number of Snomed disease hops between a question and answer can range between a continuous range of integer values. We report if restricting the number of hops to allow for more general yet precise answers improves accuracy. Similarly, in the post-sorting settings, 2 and 4, we use the feature values from the knowledge augmentation modules in addition to the LLM's own confidence scores to rank answers. Specifically, in setting 2, we sort the LLMs answerset on variations to a combination of length of disease overlap between question and answer Metamap phrases and the LLM confidence scores. In setting 5, we sort the LLMS answerset based on variations to a combination of sum of hops between question and answer Snomed disease codes, number of Snomed ancestors in the answer and LLM confidence scores. 

%SC need to expand on below method based on which method we use for the ontology filtering 
We report the accuracies for answers that address questions of question types 3 from Tab. \ref{tab:questiontypes}, that use these knowledge augmentations in the results section (Sec. \ref{sec:results}). We have written functions that use the NLTK toolkit in our evaluation submodule to generate standard, natural language processing (NLP), accuracy scores like F1, precision, recall, and BLEU. Overall, the integration of a semantic mapping tool helps us enhance the capabilities of BERT LLMs for more precise and better semantic matches via knowledge-driven methods.

\section{True Label Annotations for Guideline Questions}
\label{sec:annotations:appendix}
% \todotip[SC]{add a disclaimer about these annotations does not correspond to ADA guidance. And mention our total number here and $70$ in the main text.}
We generated annotations for creating the gold standard dataset for feature importance questions (question type 3 from question types supported by our QA approach, see Tab. \ref{tab:questiontypes}), whose results are presented in the main manuscript of the paper. These questions included disease feature importances of diagnostic value. The annotations were done by the first author by reading the ADA 2021 CPG and looking for answers to these questions. We also ran our annotations by a medical expert on our team, who is also a co-author on this paper, to verify if they were clinically meaningful. The medical expert validated $47$ questions out of the $71$ questions annotated by the first author and we report results on this expert validated set in Sec. \ref{sec:result_settings:appendix}. To the best of our knowledge, we are the first to report a dataset of questions with annotated answers on the ADA 2021 CPG. It is to be noted that these annotations are not endorsed by the ADA. Nevertheless, we hope that our annotations can serve as a valuable resource for academic advances in the clinical informatics community. 

\section{Results}
\label{sec:result_settings:appendix}
Here we present additional material to support the results of our risk prediction and question-answering module described in Sec. \ref{sec:guidelineqaresults}. 

\subsection{Model Performance: Risk-Prediction Model and Post-Hoc Explainers}
% \todotip[SC]{Move this to appendix. Add to the effect that we choose MLP, as our risk prediction model and derived the important features using SHAP.}
% \todotip[PC]{Lead}
We present the performance of the risk prediction models in Tab. ~\ref{tab:resultsCKD}. As the table shows, while GRU performs the best overall, depending on the use case, we may want to prefer other models. For the purposes of this paper, we chose MLP as our risk prediction model to benefit from the higher recall (such that the probability of false negatives is low) and high brier-score (to allow a more natural interpretation of our model outputs for clinicians), while still achieving an acceptable level of overall performance (AUC-ROC = $0.59$).

\begin{table}[ht!]
  \caption{Results of CKD risk from different prediction models}
  \centering
  %\todotip[PC]{add CI}
  \label{tab:resultsCKD}
  \begin{tabular}{lccccc}
    \toprule
    Method & Precision & Recall & AUC-ROC & AUC-PRC & Brier\\
    \midrule
    %One-class SVM & & \\
    LR    &  0.333 & 0.023  & 0.582 & 0.215 &  0.127 \\
    MLP   &  0.139 &  {\bf 0.977} & 0.587 & 0.224 & {\bf 0.621}  \\
    LSTM  &  {\bf 0.242} &  0.442 & {\bf 0.678} & 0.263 & 0.208  \\
    GRU   &  0.240  &  {\bf 0.605} & 0.677 & {\bf 0.311} & 0.220   \\
  \bottomrule
\end{tabular}
\end{table}

\begin{figure*}[!htbp]
    \centering
    % \begin{subfigure}{0.74\textwidth}
    %   \includegraphics[width=\linewidth]{./images/shap_bar_proto_CKD_w360_v1.png}
    % \caption{Feature importance}
    % \label{fig:shap:bar_proto}    
    % \end{subfigure}
    % \hspace{-1.5em}
    % \begin{subfigure}{0.257\textwidth}
    %  \adjincludegraphics[width=\linewidth,trim={{.65\width} 0 0  0},clip]{./images/shap_summary_proto_CKD_w360_v1.png}
    %   %\includegraphics[trim=10cm 0 0 0,clip=true,totalwidth=0.9\linewidth]{./images/shap_summary_proto_{\ckd}_w360_v1.png}
    %     \caption{Feature Impact}
    %     \label{fig:shap:violin_proto}
    % \end{subfigure}
    \includegraphics[width=\linewidth]{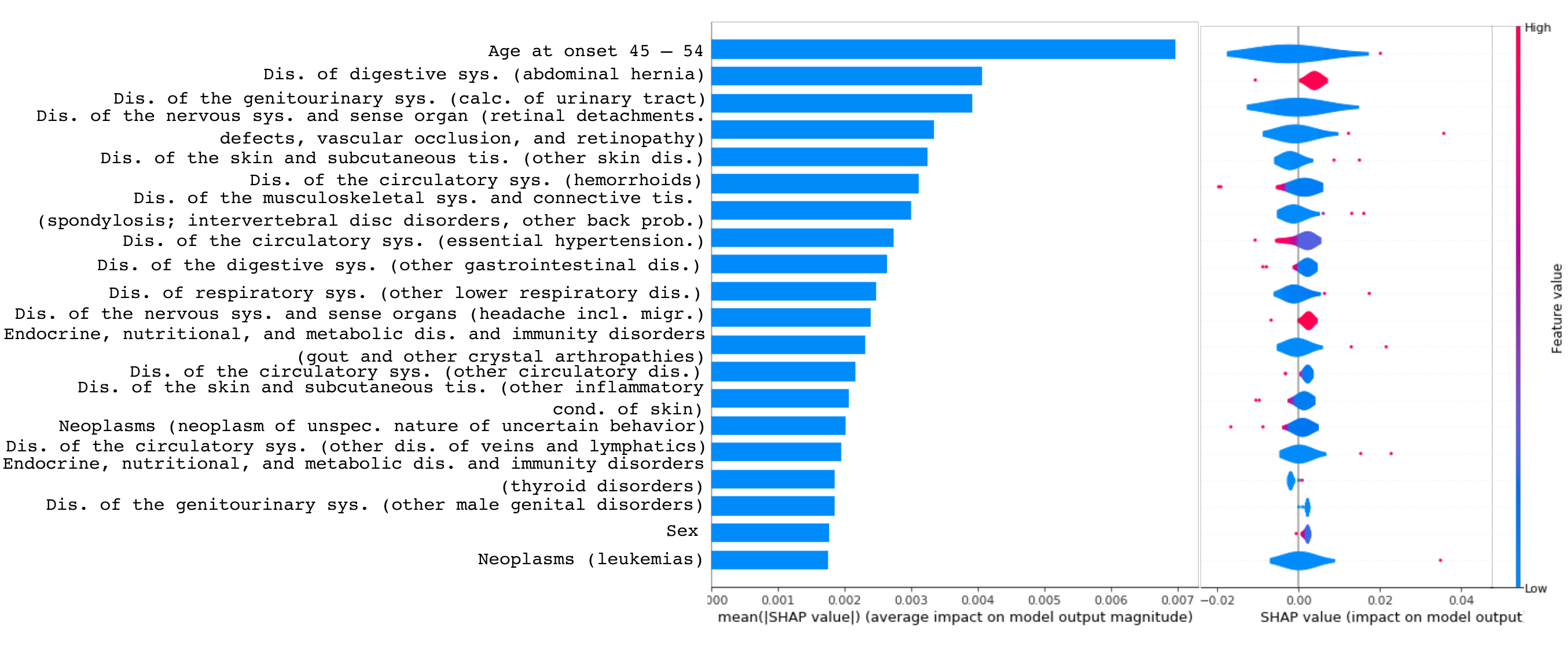}
    \vspace{-2em}
        \caption{Feature importance for {\ckd} prediction among {\numproto} prototypical patients using SHAP (left), showing absolute importance, and (right) showing feature impact on model prediction w.r.t. presence/absence of features}
        \label{fig:shap:summary_proto}
\end{figure*}

Feature importance for risk factors found by algorithmic explainers can be further contextualized, as mentioned previously.
The left hand column of Fig.~\ref{fig:shap:summary_proto} shows the top $20$ features for the set of {\numproto} prototypical patients under investigation. These prototypical patients are all found to be at high-risk for {\ckd} and hence, would be interesting to clinicians within the scope of this {\dm} and {\ckd} use case. %(\textbf{Q. 1}). 
For these prototypical patients, we present aggregated feature importance, as seen in Tab. \ref{tab:protosummary_appendix}, to account for HIPAA restrictions.  We can see that demographic features, such as age and the presence of other disorders, such as `other skin disorders,' were found to be important for the {\ckd} risk prediction. 
The right side of Fig. ~\ref{fig:shap:summary_proto} shows an alternate view of the same, providing a view into the spread of individual importance. From this deeper view, we can see that features such as `calculus of urinary tract' could be the most important drivers of risk for some patients. Such results further support our need to personalize features found to be important for the risk predictions. 
% contextualization

While insights about the importance of such features are helpful, such clinical and patho-physiological features may need further contextualization for clinicians. Our structured feedback sessions found that clinicians found the contextual explanations we support around these features helpful, and we cover some of this next.

As seen in Tab. ~\ref{tab:expertresults-native} and Tab. ~\ref{tab:expertresults-kmidiseases}, we provide results numbers on a small set of expert validated answers from our guideline annotations of $12$ questions and $47$ candidate answers. We are considering methods like weak supervision to increase the expert validation coverage of our annotations. We find that generally the results on the expert validated answers (Tab.~\ref{tab:expertresults-native} and ~\ref{tab:expertresults-kmidiseases}) follow the trend of accuracy values in the larger annotation set, in that the precision is highest in the pre-filtering by number of Snomed disease hop settings (setting 4) - $0.29$ and that the recall and bleu is high in post-filtering by number of disease overlaps between question and answer (SciBERT + KA in Tab. \ref{tab:expertresults-kmidiseases}), setting 3 - $0.6$ and $0.2$ respectively. However, contrary to the larger set of results the F1 are highest in both the vanilla BERT and BioBERT-BioASQ + KA settings, $0.22$. 

\begin{table}[!htbp]
\caption{Performance of Guideline QA with different language model approaches reported at mean average precision (map), F1 and recall at top-10 answers and precision at top-1 and top-5 for $12$ expert validated questions. Best and second-best values for each column is highlighted in \textcolor{green}{Green} and \textcolor{blue}{Blue} color, respectively. Language model (e.g. BERT) suffixed with KA represents the corresponding knowledge augmented model (e.g. BERT-KA).}
\label{tab:expertresults-native}
\small
\centering
\small
\begin{tabular}{lrrrrrr} 
{} & {bleu} & {P@1} & {P@5} & {map} & {f1} & {recall} \\
{model} & {} & {} & {} & {} & {} & {} \\
BERT & \color{green} 0.155 & \color{blue} 0.363 & \color{green} 0.262 & \color{green} 0.267 & \color{blue} 0.224 & \color{blue} 0.365 \\
BioBERT & 0.121 & 0.296 & 0.200 & 0.222 & 0.186 & 0.342 \\
BioBERT-BioASQ & 0.131 & 0.259 & 0.200 & 0.205 & 0.192 & 0.363 \\
BioClinicalBERT-ADR & 0.112 & 0.227 & 0.178 & 0.179 & 0.171 & 0.351 \\
SciBERT & \color{blue} 0.153 & \color{green} 0.366 & \color{blue} 0.216 & \color{blue} 0.244 & \color{green} 0.235 & \color{green} 0.463 \\
\end{tabular}
\end{table}

\begin{table}[!htbp]
\caption{Performance of Guideline QA with different language model approaches + knowledge augmentations reported at mean average precision (map), F1 and recall at top-10 answers and precision at top-1 and top-5 for $12$ expert validated questions. Best and second-best values for each column are highlighted in \textcolor{green}{Green} and \textcolor{blue}{Blue} color, respectively. Language model (e.g. BERT) suffixed with KA represents the corresponding knowledge augmented model (e.g. BERT-KA).}
\label{tab:expertresults-kmidiseases}
\small
\centering
\small
\begin{tabular}{lrrrrrr}
{} & {bleu} & {P@1} & {P@5} & {map} & {f1} & {recall} \\
{model} & {} & {} & {} & {} & {} & {} \\
BERT-KA & 0.021 & 0.296 & \color{green} 0.296 & \color{green} 0.296 & 0.127 & 0.081 \\
BioBERT-KA & 0.143 & \color{green} 0.440 & 0.215 & 0.258 & 0.222 & 0.319 \\
BioBERT-BioASQ-KA & \color{blue} 0.147 & \color{blue} 0.366 & \color{blue} 0.249 & 0.272 & \color{blue} 0.227 & 0.335 \\
BioClinicalBERT-ADR-KA & 0.123 & 0.321 & 0.209 & 0.221 & 0.201 & \color{blue} 0.384 \\
SciBERT-KA & \color{green} 0.201 & 0.356 & 0.209 & \color{blue} 0.284 & \color{green} 0.297 & \color{green} 0.600 \\
\end{tabular}
\end{table}

\begin{figure*}[!htbp]
    \centering
    \includegraphics[width=\linewidth]{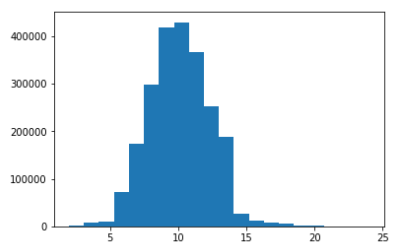}
        \caption{Distribution of number of hops between Snomed disease pairs from questions and candidate guideline answers. As can be seen most disease pairs are between $5$ - $15$ hops apart with $20$ being the maximum number of hops.}
        \label{fig:snomedhops}
\end{figure*}

In Fig. \ref{fig:snomedhops}, we also report the distribution of the number of hops between disease code pairs found in the question and candidate answers to provide an idea of how far or close the questions are to sentences within the guidelines. As can be seen, most question and answer pairs are between $5$ - $15$ hops away. We find that the results are best when we filter answer codes less than $3 - 6$ hops away from the question disease codes. 

%SC From Ching-Hua we do not need to release code at this time
% \section{Code Resources}
% \label{sec:codesetup:appendix}
% Attach guideline QA code as supplementary material. 
% %SC It is fine if we don't submit this code in this pass
%% If you have bibdatabase file and want bibtex to generate the
%% bibitems, please use
%%

\section{Risk Prediction Dashboard Description}
\label{sec:dashboard:appendix}

\begin{figure*}[!htbp]
    \centering
    \includegraphics[width=\linewidth]{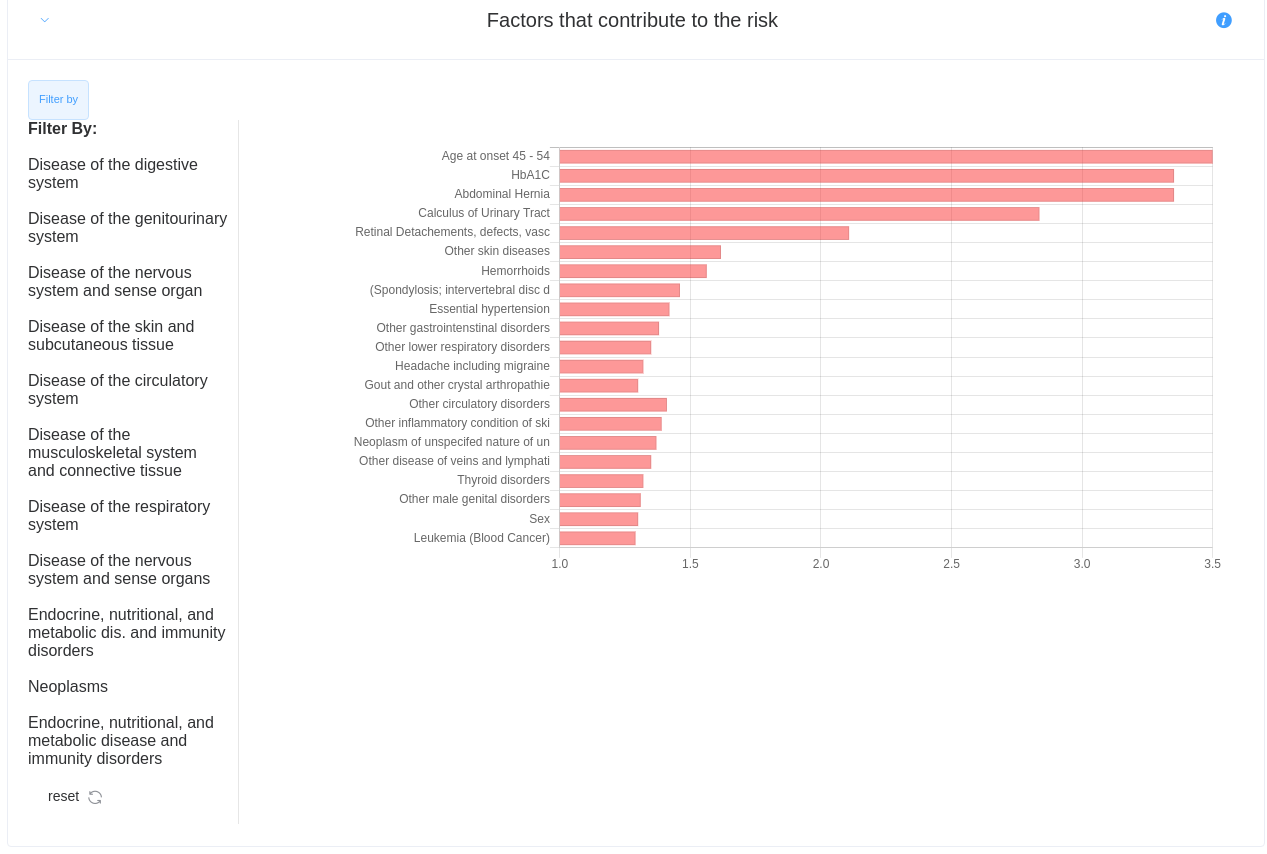}
        \caption{The feature importances pane of our risk prediction dashboard shows in order of importance the features that contributed to the patient's predicted {\ckd} risk. The diagnostic features can also be filtered by their higher-level disease groupings and can be selected from the filter by column seen on the left of this figure.}
        \label{fig:featurespane}
\end{figure*}

\begin{figure*}[!htbp]
    \centering
    \includegraphics[width=\linewidth]{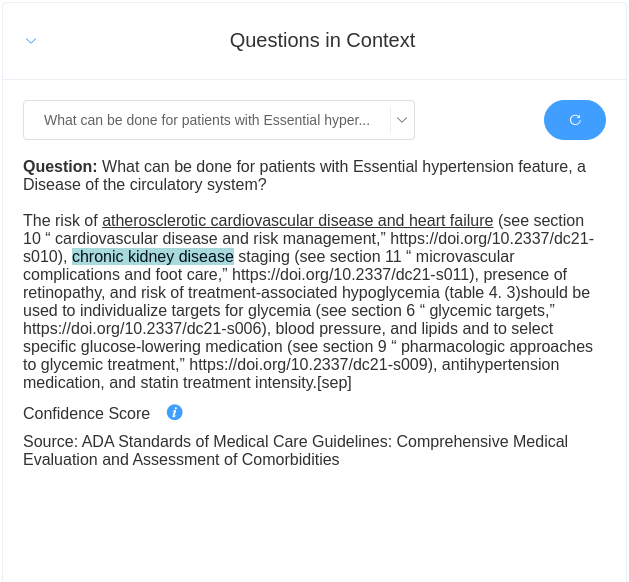}
        \caption{Here is the questions in context pane, which has a list dropdown option as seen on the top of this figure, where clinicians can browse through the question list we support for the patient being shown. These questions span the different question types we support, and the length of the question list is variable depending on how many diagnostic features contributed to the patient's predicted risk. Also seen here is the detail we support for each question and their answer, including provenance details like the confidence score and the data source for the predicted answer.}
        \label{fig:questionsincontextpane}
\end{figure*}

We provide additional views of the different panes in our risk prediction dashboard that hosts our supported contextual explanations alongside the different entities that we contextualize, the patient, their risk prediction and the important features found to contribute to the risk prediction. We also support interactions between each of these panes which can help clinicians easily find the content that contextualizes the entities. These interactions or brushing capabilities include (see Fig. \ref{fig:riskpredictionpane}, \ref{fig:featurespane}, \ref{fig:questionsincontextpane}): 

\begin{figure*}[!htbp]
    \centering
    \includegraphics[width=\linewidth]{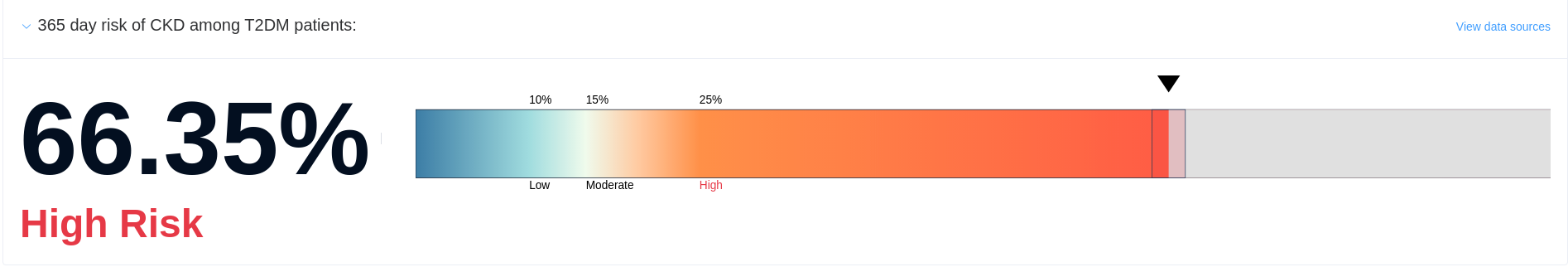}
        \caption{The risk prediction pane of our risk prediction dashboard, wherein the risk score is displayed alongside a severity of the score on a threshold scale. }
        \label{fig:riskpredictionpane}
\end{figure*}

\begin{itemize}
    \item Clicking on the patient details pane brings up questions in the questions in context pane asking about the patient's diabetes state 
    \item When a month is chosen on the history timeline, questions about a commonly accepted diabetes indicator, Hemoglobin A1C (HbA1C), are brought up or filtered in the questions in context pane
    \item Clicking anywhere in the risk prediction pane brings up questions about the patient's predicted risk
    \item In the feature importances pane, features can be filtered by the selection of their corresponding, higher-level disease grouping 
    \item Clicking on a feature in the feature importance chart brings up questions about the feature 
    \item Finally, clicking on a higher-level disease grouping also brings up questions about the disease grouping
\end{itemize}

\end{document}